\documentclass[letterpaper]{article} 
\usepackage{aaai25}  
\usepackage{times}  
\usepackage{helvet}  
\usepackage{courier}  
\usepackage[hyphens]{url}  
\usepackage{graphicx} 
\usepackage{natbib}  
\usepackage{caption} 
\usepackage{algorithm}
\usepackage{algorithmic}
\usepackage{amsmath}
\usepackage{amsfonts}
\usepackage{multirow}
\usepackage{multicol}
\usepackage{dsfont}
\usepackage{booktabs}
\usepackage{colortbl}
\usepackage{pifont}
\usepackage{adjustbox}

\usepackage[dvipsnames]{xcolor}
\definecolor{lightblue}{rgb}{0.84,0.91,0.97}
\definecolor{lightgray}{rgb}{0.93,0.93,0.93}

\definecolor{lightgray}{rgb}{0.93,0.93,0.93}
\definecolor{lightyellow}{rgb}{1.0,0.97,0.88}

\newcommand{\q}[1]{`#1'}
\newcommand{\dq}[1]{``#1''}
\newcommand{\ie}{\textit{i}.\textit{e}., }
\newcommand{\eg}{\textit{e}.\textit{g}., }

\usepackage{newfloat}
\usepackage{listings}
\DeclareCaptionStyle{ruled}{labelfont=normalfont,labelsep=colon,strut=off} 
\floatstyle{ruled}
\newfloat{listing}{tb}{lst}{}
\floatname{listing}{Listing}
\title{VidChain: Chain-of-Tasks with Metric-based \\ Direct Preference Optimization for Dense Video Captioning}

\author{
    Ji Soo Lee\equalcontrib,
    Jongha Kim\equalcontrib,
    Jeehye Na,
    Jinyoung Park,
    Hyunwoo J. Kim\thanks{Corresponding author.} \\
}
\affiliations{
    Department of Computer Science and Engineering, Korea University \hspace{0.4cm} \\
    \{simplewhite9, jonghakim, sonicdog00, lpmn678, hyunwoojkim\}@korea.ac.kr
}

\usepackage{bibentry}

\begin{document}
\maketitle

\begin{abstract}
Despite the advancements of Video Large Language Models (VideoLLMs) in various tasks, they struggle with fine-grained temporal understanding, such as Dense Video Captioning (DVC).
DVC is a complicated task of describing all events within a video while also temporally localizing them, which integrates multiple fine-grained tasks, including video segmentation, video captioning, and temporal video grounding.
Previous VideoLLMs attempt to solve DVC in a single step, failing to utilize their reasoning capability.
Moreover, previous training objectives for VideoLLMs do not fully reflect the evaluation metrics, therefore not providing supervision directly aligned to target tasks.
To address such a problem, we propose a novel framework named VidChain comprised of Chain-of-Tasks (CoTasks) and Metric-based Direct Preference Optimization (M-DPO).
CoTasks decompose a complex task into a sequence of sub-tasks, allowing VideoLLMs to leverage their reasoning capabilities more effectively.
M-DPO aligns a VideoLLM with evaluation metrics, providing fine-grained supervision to each task that is well-aligned with metrics.
Applied to two different VideoLLMs, VidChain consistently improves their fine-grained video understanding, thereby outperforming previous VideoLLMs on two different DVC benchmarks and also on the temporal video grounding task.
Code is available at \url{https://github.com/mlvlab/VidChain}.
\end{abstract}

\section{Introduction}
With the rapid advancement of Large Language Models (LLMs), numerous studies~\cite{liu2023visual,dai2023instructblip,liu2024improved} have incorporated LLMs into video understanding tasks, leading to the emergence of Video Large Language Models (VideoLLMs).
These VideoLLMs~\cite{li2023videochat,zhang2023video,maaz2023video} have demonstrated strong performance in various tasks such as video question answering and video captioning, showcasing their ability to understand and utilize visual information. 
Despite their success, recent studies~\cite{ren2024timechat, huang2024vtimellm, qian2024momentor} have revealed that VideoLLMs exhibit unsatisfactory performance when it comes to fine-grained temporal video understanding, which often require \textit{multiple} video-related sub-tasks given an untrimmed video. 

We observe that VideoLLMs fall short of fine-grained temporal video understanding especially in Dense Video Captioning (DVC) due to two key reasons.
First, the conventional practice in DVC of VideoLLMs employs one-step reasoning, which is known to be inferior to multi-step reasoning for complex tasks.
In particular, existing VideoLLMs address DVC by predicting descriptions and timestamps of all events via a single-step generation.
Second, the gap between training objectives (\textit{e.g.}, next-token prediction) and evaluation metrics for DVC (\textit{e.g.}, SODA) often leads to suboptimal performance.
The next-token prediction does not fully reflect the complex evaluation protocol which involves diverse metrics such as SODA, METEOR, and IoU.

To tackle the aforementioned issues, we introduce a novel framework, VidChain that enhances VideoLLMs' fine-grained temporal video understanding, comprised of Chain-of-Tasks (CoTasks), and Metric-based Direct Preference Optimization (M-DPO).
First, we present CoTasks that decompose the objective of the challenging task into a sequence of sub-task objectives.
This simple decomposition enables the model to elicit its strong reasoning capability on DVC.
It eases the challenge of the complex task by solving only one sub-task at each step and enhances its capability of fine-grained temporal video understanding.
Second, to further align VideoLLM with the evaluation metrics of DVC, we present M-DPO which learns the \textit{metric preference}, a preference based on the evaluation metric such as SODA, of each sub-task that composes DVC.
Following the insight from DPO~\cite{rafailov2024direct}, which aligns LLM with human preferences, we adopt a similar approach yet we align VideoLLMs specifically with the metric preferences.

Interestingly, we observe that this simple adaptation of evaluation metrics provides two advantages: (1) it reduces the reliance on human annotators being cost-efficient. 
(2) metric evaluations expand beyond the standard binary decision dataset where the labels are \textit{continuous} \textit{e.g.}, 10.0, 8.5, rather than \textit{discrete} \textit{e.g.}, win or lose. 
Moreover, we take account of the sequential sub-task prediction in CoTasks, where we supervise metric preferences on the \textit{final} response of the model as well as on the \textit{intermediate} sub-tasks that allow for more fine-grained supervision.
Overall, our M-DPO is a novel method that reflects continuous characteristics of the metric-based evaluations into learning, and also provides intermediate task-specific supervision, further enhancing fine-grained video understanding of VideoLLMs.
We evaluate our VidChain on two benchmarks-Activitynet Captions and YouCook2 for the challenging DVC task, and Activitynet Captions for temporal video grounding (TVG).

In sum, our contributions are three-fold:
\begin{itemize}
    \item We propose Chain-of-Tasks (CoTasks) that decomposes a complicated task into a sequence of sub-tasks, enabling the VideoLLM to elicit its strong reasoning capability to address the challenging task of DVC.
    \item We present Metric-based Direct Preference Optimization (M-DPO) that aligns VideoLLM with evaluation metrics for multiple fine-grained video understanding tasks, providing supervision targeted to each task.
    \item Our novel framework, VidChain comprising of CoTasks and M-DPO, is generally applicable to LLM-based models which consistently improves performances when applied to baseline models.
\end{itemize}

\section{Related Works}
\paragraph{Dense Video Captioning.}
A fine-grained temporal understanding task for videos, Dense Video Captioning (DVC), expects the model to caption the events in the long untrimmed video, and temporally localize them.
An early approaches to DVC include PDVC~\cite{wang2021end}, which introduced a variant of the transformer-based DETR~\cite{zhu2020deformable} to predict the event captions and their timestamps.
Subsequent works~\cite{yang2023vid2seq, yang2024vidchapters, kim2024you} explored incorporating additional data or modalities as external knowledge, such as transcribed speech or external caption to enhance performance.
Another study~\cite{zhou2024streaming} addressed online dense video captioning, processing videos frame by frame instead of analyzing them as a whole.
More recently, the emergence of Video Large Language Models has inspired efforts to evaluate these models' fine-grained video understanding capabilities with the task of DVC.
However, an effective strategy to enhance VideoLLMs on DVC remains underexplored.

\paragraph{Video Large Language Models.}
Recently, multiple works~\cite{liu2023visual,dai2023instructblip,liu2024improved, chen2024internvl, ye2024mplug, ko2023open,ko2023large} incorporating Large Language Models (LLMs) for vision-language tasks have been proposed.
Following those models' successes, several Video Large Language Models (VideoLLMs) have been proposed~\cite{li2023videochat,zhang2023video,maaz2023video, zhu2023languagebind, lin2023video, li2024mvbench}.
Despite the remarkable performance of VideoLLMs in tasks requiring a holistic understanding of a video (\textit{e.g.}, video-level question-answering or captioning), they often fall short in fine-grained video understanding.
For instance, they often suffer in temporal grounding tasks~\cite{krishna2017dense} or dense video captioning tasks~\cite{krishna2017dense,zhou2018towards}, where diverse fine-grained video understanding capabilities are required.
Thus, multiple works~\cite{ren2024timechat,huang2024vtimellm,qian2024momentor} have tried incorporating fine-grained information into VideoLLMs to address the problem.
In this study, we propose decomposing a complicated task of DVC into simpler sub-tasks and providing supervision aligned with the desired capability, thereby enhancing VideoLLMs' capability in fine-grained understanding.

\paragraph{Direct Preference Optimization.}
To align LLM outputs with human preferences, reinforcement learning from human feedback (RLHF)~\cite{christiano2017deep,ouyang2022training} has been proposed, which maximizes the likelihood gap between the preferred and unpreferred generation results.
Direct preference optimization (DPO)~\cite{rafailov2024direct} is derived to improve the inefficiency of RLHF, lifting the need for RL-based optimization and dedicated modules (\textit{i.e.}, reward model), which is applied across tasks~\cite{song2024preference,xu2024contrastive,yuan2024advancing} to inject human preferences.
Recently, some works have explored preference alignment in multimodal language models~(MLLMs) to alleviate the hallucination issue~\cite{yu2024rlhf,ahn2024tuning,gunjal2024detecting}.
However, these approaches rely on expensive models like GPT-4v~\cite{ahn2024tuning} or human annotators~\cite{yu2024rlhf}.
In this work, we adopt the idea of Step-DPO~\cite{lai2024step} to align VideoLLM on every sub-task with the desired capability in fine-grained video understanding by defining the preferred and unpreferred responses using the metric as a criterion.
Such an approach eliminates the need for extensive human labor or computation.
Also, unlike conventional binary preference datasets, ours leverages continuous preferences based on the metrics.

\begin{figure*}[ht!]
    \centering
    \includegraphics[width=1.0\textwidth]{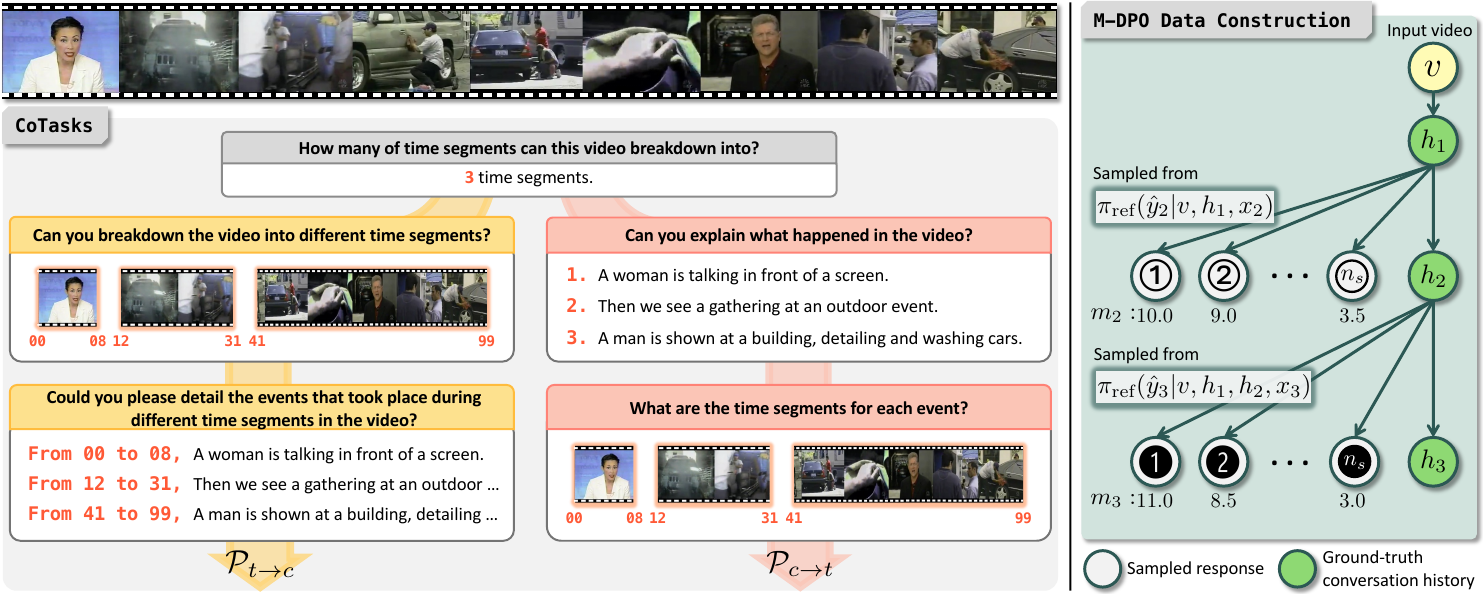}
    \begin{minipage}[b]{0.352\textwidth}
        \centering
        \caption*{(a)}
    \end{minipage}
    \begin{minipage}[b]{0.367\textwidth}
        \centering
        \caption*{(b)}
    \end{minipage}
    \begin{minipage}[b]{0.271\textwidth}
        \centering
        \caption*{(c)}
    \end{minipage}
    \caption{\textbf{Illustration of our CoTasks approach (left) and data construction process for M-DPO (right).}
    The left figure depicts the CoTasks approach of VidChain, which decomposes DVC into a sequence of sub-tasks in two different reasoning paths.
    After predicting the number of events, timestamp prediction and caption generation are done in path $\mathcal{P}_{t \rightarrow c}$ as shown in (a), while the order of two tasks is interchanged in path $\mathcal{P}_{c \rightarrow t}$ as in (b).
    The right figure (c) represents the data construction of M-DPO, where we sample $n_s$ response of $\hat{y}_3$ (filled black circles) as well as the intermediate sub-task response $\hat{y}_2$ (hollow black circles) of CoTask for the given video $v$.
    The $m_2$ and $m_3$ denote the task-specific evaluation metric values, \textit{e.g.}, $\text{SODA}_c$, METEOR, IoU, of each sampled response.
    }
    \label{fig:main_fig}
\end{figure*}

\section{Method}
In this section, we first provide a brief overview of Dense Video Captioning (DVC) and Direct Preference Optimization (DPO) in Sec.~\ref{preliminaries}.
Then, we propose a Chain-of-Tasks (CoTasks) approach which eases the challenge of DVC by decomposing the task into a sequence of sub-tasks (Sec.~\ref{sec:CoTasks}). 
We then present a Metric-based DPO (M-DPO), which further aligns VideoLLMs with evaluation metrics for sub-tasks (\textit{e.g.,} METEOR, IoU, SODA) to provide more fine-grained supervision (Sec.~\ref{Progressive-DPO}).
Comprised of CoTasks and M-DPO, we propose a novel framework named VidChain that enhances VideoLLMs' fine-grained temporal video understanding capability.

\subsection{Preliminaries}
\label{preliminaries}
\paragraph{Dense Video Captioning.}
An challenging task of Dense Video Captioning requires the model to not only describe all events within the long untrimmed video but also temporally localize each event in time.
Given a video $v$, the goal of DVC is to maximize the probability $p(c, t, n | v)$, where $n$ denotes the number of events in the video, $c$ denotes a set of event captions, and $t$ denotes a set of event timestamps represented with start and end time boundaries for each.
The key challenge is that the model must predict the three components ($c$, $t$, $n$) for all events given an untrimmed video $v$, which requires a comprehensive fine-grained understanding regarding multiple video-related tasks: video segmentation, video captioning, and temporal video grounding. 
The video segmentation task aims to predict the number of event sequences the video breaks down into.
Video captioning focuses on describing the events in the video.
The temporal video grounding task aims to identify the timestamps of an event given its event description.
Typically, VideoLLMs address DVC by predicting ($c$, $t$, $n$) in a single step, which imposes more challenge on the task.

\paragraph{Direct Preference Optimization.}
Direct Preference Optimization (DPO)~\cite{rafailov2024direct} aligns Large Language Models' output with human preferences.
Often, the alignment involves further finetuning a supervised finetuned model. 
For optimization, DPO adopts a pairwise preference dataset $\mathcal{D}_{\text{DPO}}$, where each sample comprises a pair of preferred and dispreferred responses.
To construct the dataset, several responses $\hat{y}$ are first sampled from the reference model  $\pi_{\text{ref}}$ given a prompt $x$.
Then, those responses are annotated according to the human preferences by comparing sampled responses in a pair-wise manner where $\hat{y}^{w}$ and $\hat{y}^{l}$ denote the preferred and dispreferred response respectively, \textit{i.e.}, $\hat{y}^{w} \succ \hat{y}^{l} | x$.
Hence, the objective of DPO, $\mathcal{L}_{\text{DPO}}$, is formally defined as follows:
\begin{equation}
\fontsize{9.5pt}{10pt}
\begin{split}
\mathcal{L}_{\text{DPO}}  (\pi_{\theta};  \pi_{\text{ref}}) & = - \mathbb{E}_{(v, x, \hat{y}^{w}, \hat{y}^{l}) \sim \mathcal{D}_{\text{DPO}}} \\
& \bigg[\log \sigma \Big(\beta \cdot r(\hat{y}^w;v,x) - \beta \cdot r(\hat{y}^l;v,x)\Big)\bigg], 
\end{split}
\end{equation}
where $r(y;v, x) =  \log \frac{\pi_{\theta}\left( y |v,x\right)}{\pi_{\text{ref}}\left(y|v,x \right)}$.
This is identical to the original DPO, yet we include video $v$ to account for the video context.
$\sigma(\cdot)$ is the sigmoid function, $\pi_{\theta}$ is the model to be optimized, which is initialized to $\pi_{\text{ref}}$, and $\beta$ is a hyperparameter that controls the distribution disparity of $\pi_{\theta}$ from the reference model $\pi_{\text{ref}}$.
Overall, the model is trained to increase the likelihood of the preferred responses relative to that of dispreferred responses.

\subsection{Chain-of-Tasks}
\label{sec:CoTasks}
To address the lack of fine-grained temporal understanding of VideoLLMs, especially in DVC that encompasses multiple video-related tasks, we propose a novel approach Chain-of-Tasks.
Most prior works~\cite{ren2024timechat, qian2024momentor, huang2024vtimellm, yang2023vid2seq} address DVC by directly predicting ($c, t, n$) given $v$ within a single step.
Yet, this approach imposes more challenges on the task for the VideoLLM, as it obstructs the model from leveraging its strong reasoning capability.
Hence, in CoTasks, we first decompose the objective of DVC into a series of sequential sub-task objectives.
Then we prompt each task corresponding to each objective to the VideoLLM in the form of a multi-turn QA conversation.
Such an approach eases the challenge of DVC by solving only one sub-task at each turn and further enhances a VideoLLM's capability in fine-grained temporal video understanding.

\paragraph{Objective Decomposition.}
Objective of DVC can be decomposed in two different reasoning paths, $\mathcal{P}_{t \rightarrow c}$ and $\mathcal{P}_{c \rightarrow t}$:
\begin{align}
 p(c, t, n|v)
&= p(c|v, n, t) p(t|v, n) p(n|v) \ \ \ \ \ (\mathcal{P}_{t \rightarrow c})\label{eq:timefirst} \\
&= p(t|v, n, c) p(c|v, n) p(n|v) \ \ \ \ \ (\mathcal{P}_{c \rightarrow t})\label{eq:capfirst}. 
\end{align}
In the case of $\mathcal{P}_{t \rightarrow c}$ in Eq.~\eqref{eq:timefirst}, the prediction of $(c, t, n)$ given a video $v$ breaks down into three sequential tasks.
First, $p(n|v)$ represents the task of predicting the number of events in the video, while the following $p(t|v, n)$ represents the task of the timestamp prediction given the total number of events $n$, and the final $p(c|v, n, t)$ indicates caption generation for each event given the video with its $t$ and $n$.
The other path $\mathcal{P}_{c \rightarrow t}$ in  Eq.~\eqref{eq:capfirst} is also similarly defined, except that the order of the caption generation and the timestamp prediction tasks are interchanged.
Based on this decomposition, we cast the task of DVC as a multi-turn prediction, where the model sequentially solves different tasks at each turn to tackle the challenging task.
An example of our multi-turn approach, namely CoTasks, is illustrated in Fig.~\ref{tab:main}.(a) and~\ref{tab:main}.(b), each corresponding to $\mathcal{P}_{t \rightarrow c}$, and $\mathcal{P}_{c \rightarrow t}$.

\paragraph{Training data construction for CoTasks.}
$\mathcal{D}_{\text{CT}}$ is a multi-turn conversation dataset used for training VideoLLMs to reason in a CoTasks manner.
We build CoTasks samples using the original DVC dataset (\eg ActivityNet or YouCook2), by converting the original single-turn conversation samples into multi-turn CoTasks samples of both $\mathcal{P}_{t \rightarrow c}$ and $\mathcal{P}_{c \rightarrow t}$ types.
We construct 10K and 1K samples for AcitivtyNet and YouCook respectively for each path using the pre-defined templates, where the templates are provided in the supplement.
Note we refer to each of the two types of dataset as $\mathcal{D}_{t \rightarrow c}$ and $\mathcal{D}_{c \rightarrow t}$, respectively.
Then, we combine our obtained dataset with the DVC QA pairs and dialogues following VTimeLLM~\cite{huang2024vtimellm}.
Note that we adopt the full benchmark dataset unlike VTimeLLM, which only uses a selected subset for training.
This results in $\mathcal{D}_{\text{CT}}$ of size of 50K for ActivityNet and 6K for YouCook2.
Overall, we use $\mathcal{D}_{\text{CT}}$ to finetune VideoLLMs, enhancing their performance on fine-grained video understanding tasks, including DVC and its sub-tasks.
Further details are in the supplement.

\paragraph{Inference pipeline of CoTasks.}
\label{subsec:CoTasks-inference}
Since $\mathcal{D}_{\text{CT}}$ includes both samples of $\mathcal{D}_{t \rightarrow c}$ and $\mathcal{D}_{c \rightarrow t}$, the VideoLLM trained with $\mathcal{D}_{\text{CT}}$ can take either reasoning path during inference to address DVC.
To encourage the model to take a certain path, we prompt the model with path-specific prompts.
For instance, for $\mathcal{P}_{c \rightarrow t}$, we prompt \dq{Can you explain what happened in the video?} to encourage the generation of event captions first, after addressing the common task for both paths, \textit{i.e.}, the number of event predictions.
In our experiments, we provide results in both inference paths.

\subsection{Metric-based Direct Preference Optimization}
\label{Progressive-DPO}
Although CoTasks enhances VideoLLMs in fine-grained video understanding tasks, the next-token prediction objective does not fully reflect the complex evaluation protocol which involves diverse metrics such as SODA, METEOR, and IoU.
Therefore, we propose a novel optimization method named Metric-based Direct Preference Optimization (M-DPO).
Inspired by DPO, which aligns a model with human preferences, M-DPO aligns the VideoLLM with metric preferences, where the evaluation metric for tasks within CoTasks is adopted as criteria to determine preferred and dispreferred responses.
This approach enables more fine-grained metric preference alignment of the VideoLLM as it not only supervises the \textit{final} response but also across the \textit{intermediate} responses within CoTasks.
In the following sections, we first describe the process of constructing a dataset used for M-DPO training, which adopts metrics as criteria.
Then, we introduce the overall training objective of M-DPO.
Finally, we present a preference gap-aware M-DPO, which is an extension of M-DPO equipped with a tailored training scheme reflecting the continuous nature of metrics.

\begin{table*}[t!]
    \centering
    \small
    \renewcommand{\arraystretch}{1.1}
    \setlength{\tabcolsep}{3mm}
    \begin{tabular}{l|c|ccc|ccc}
    \toprule
      & & \multicolumn{3}{c|}{ActivityNet} & \multicolumn{3}{c}{YouCook2}   \\ 
      & size & $\text{SODA}_c$ & METEOR & CIDEr & $\text{SODA}_c$ & METEOR & CIDEr\\
    \hline

    VideoChat ~\cite{li2023videochat} & 7B 
        & 0.9  &  0.9 & 2.2
        & -  &  - & - \\
    VideoLLaMA~\cite{zhang2023video} & 7B
        & 1.9  &  1.9 & 5.8 
        & - &  -  & -  \\
    VideoChatGPT~\cite{maaz2023video} & 7B
        & 1.9  &  2.1 & 5.8 
        & - &  -  & -    \\
    TimeChat~\cite{ren2024timechat} & 13B
        & -  &  - & - 
        & 3.4 &  -  & 11.0    \\
    VTimeLLM~\cite{huang2024vtimellm} & 13B 
        & 5.9  &  6.7 & 27.2 
        & - &  -  & -    \\
    \hline
    \hline
    VTimeLLM$^\dagger$~\cite{huang2024vtimellm}~(Baseline) & 7B
        & 5.8  &  6.8 & 27.6
        & 3.4 &  3.5  & 10.7    \\
    \rowcolor{lightblue}
    VTimeLLM + VidChain-$\mathcal{P}_{t \rightarrow c}$ \textbf{(Ours)}  & 7B   
       & 6.9 &  7.1  & 29.1
        & \textbf{4.6} &  \textbf{4.9}  & \textbf{17.6} \\

    \rowcolor{lightblue}
    VTimeLLM + VidChain-$\mathcal{P}_{c \rightarrow t}$ \textbf{(Ours)} & 7B
        & \textbf{7.2} &  \textbf{7.7}  & \textbf{34.7}
        & 4.3 &  4.5  & 16.3 \\

    \hline
    VideoLLaMA2$^\dagger$~\cite{cheng2024videollama}~(Baseline) & 7B
        & 7.2 &  7.7  & 32.9    
        & 3.3 &  3.5  & 12.3 \\
    \rowcolor{lightblue}
    VideoLLaMA2 + VidChain-$\mathcal{P}_{t \rightarrow c}$ \textbf{(Ours)}  & 7B   
        & 8.2 & 8.7  & 43.1
        & 4.6 &  5.5  & 22.3   \\

    \rowcolor{lightblue}
    VideoLLaMA2 + VidChain-$\mathcal{P}_{c \rightarrow t}$ \textbf{(Ours)} & 7B 
        & \textbf{8.8} &  \textbf{8.8}  & \textbf{43.9}    
        & \textbf{4.8} &  \textbf{5.6}  & \textbf{23.8} \\
    
    \hline
    \end{tabular}
    \caption{\textbf{Comparison of VideoLLMs on DVC.}
    Baseline+VidChain-$\mathcal{P}_{t \rightarrow c}$ and Baseline+VidChain-$\mathcal{P}_{c \rightarrow t}$ are identical models trained with $\mathcal{D}_{\text{CT}}$ which adopt two different reasoning path prompts for inference, $\mathcal{P}_{t \rightarrow c}$ and $\mathcal{P}_{c \rightarrow t}$ respectively, and size denotes LM size.
    See Sec.~\ref{subsec:CoTasks-inference} for more detail.
    $\dagger$ denotes reproduced results adopting the full benchmark dataset.
    }
    \label{tab:main}
\end{table*}

\begin{table}[t!]
    \centering
    \renewcommand{\arraystretch}{1.1}
    \begin{adjustbox}{width=1.0\columnwidth}
    \begin{tabular}{l|c|cccc}
    \toprule
      & size & R@0.3 & R@0.5 & R@0.7 & mIoU  \\
    \hline

    VideoChat & 7B & 8.8  &  3.7 & 1.5 & 7.2 \\
    VideoLLaMA& 7B & 6.9 &  2.1  & 0.8  & 6.5 \\
    VideoChatGPT& 7B & 26.4 &  13.6  & 6.1  & 18.9 \\
    VTimeLLM & 13B & 44.8 &  29.5  & 14.2  & 31.4 \\
    \hline
    \hline
    VTimeLLM~(Baseline) & 7B & 44.0 &  27.8  & 14.3  & 30.4 \\
    $\text{VTimeLLM}^{*}$~(TVG only) & 7B & 55.8 &  35.0  & 18.9  & 37.9 \\
    \rowcolor{lightblue}
    VTimeLLM + VidChain \textbf{(Ours)}  & 7B & \multirow{1}{*}{\textbf{61.4}} &  \multirow{1}{*}{\textbf{43.8}}  & \multirow{1}{*}{\textbf{25.7}}  & \multirow{1}{*}{\textbf{43.5}} \\
    \hline
    VideoLLaMA2~(Baseline) & 7B & 49.4 &  26.8  & 15.0  & 33.9 \\
    $\text{VideoLLaMA2}^{*}$~(TVG only) & 7B & 59.9 &  41.5  & 22.5  & 42.2 \\
    \rowcolor{lightblue}
    VideoLLaMA2 + VidChain \textbf{(Ours)}  & 7B & \multirow{1}{*}{\textbf{63.3}} &  \multirow{1}{*}{\textbf{44.8}}  & \multirow{1}{*}{\textbf{25.2}}  & \multirow{1}{*}{\textbf{44.1}}  \\
    \hline
    \end{tabular}
    \end{adjustbox}
    \caption{\textbf{Comparison of VideoLLMs on TVG.}
    We adopt the task-specific prompt for TVG instead of 
    two different inference prompts (\textit{i.e.}, $P_{t\rightarrow c}$, and $P_{c \rightarrow t}$) specifically defined for DVC, since they are not applicable to TVG.
     * indicates the model further trained on the corresponding TVG dataset.
    }
    \label{tab:main-tvg}
\end{table}

\paragraph{Training data construction for M-DPO.}
$\mathcal{D}_{\text{M-DPO}}$ is a preference dataset used for further aligning a VideoLLM with metric preferences, where each sample includes a pair of preferred and dispreferred responses for each specific task in the CoTask approach.
To obtain a preference pair, $n_s$ number of responses are first sampled for each intermediate $k$-th task given a video $v$, prompt $x_k$ for $k$-th task, and the conversation history $h_{< k}$ that consists of prompts $x_{< k}$ and ground-truth responses $y_{< k}$ of previous $k-1$ tasks.
Starting from $k = 2$, a single sampled response $\hat{y}_k$ is represented as:
\begin{equation}
        \hat{y}_{k} \sim \pi_{\text{ref}}(\hat{y}_{k}| v, h_{<k}, x_k),
    \label{eq:dpo_data_sampling}
\end{equation}
where the reference model $\pi_\text{ref}$ is a VideoLLM trained with $\mathcal{D}_{\text{CT}}$ for CoTasks, and $h_{<2}$ is the ground-truth conversation history consisting of the prompt $x_1$ and ground-truth response $y_1$ corresponding to $p(n|v)$.
For instance, in $\mathcal{P}_{t \rightarrow c}$ path of CoTasks, $\hat{y}_{3}$ is a response sampled from $\pi_{\text{ref}}(\hat{y}_{3}| v, h_{<3}, x_3)$, which models $p(c|v, n, t)$.  
Similarly, $\hat{y}_{2}$ is a response modeling $p(t|v, n)$.
With the $n_s$ sampled responses for each task, $\binom{n_s}{2}$ pairs of responses are obtained.
Then, for each pair, a response $\hat{y}_k$ with a higher evaluation metric value $m_{k} = \mathcal{M}_{k}(\hat{y}_{k}, y_{k})$ is set as the preferred response $\hat{y}^w_{k}$, and the other response is set as the dispreferred response $\hat{y}^l_{k}$, where $y_k$ denotes ground-truth response of $k$-th task, and $\mathcal{M}_{k}$ denotes a metric corresponding to $k$-th task (\ie METEOR, IoU, $\text{SODA}_c)$.
In other words, $\hat{y}^w_{k} \succ \hat{y}^l_{k} | v, h_{< k}, x_k$, given $m_k^w > m_k^l$ .
The illustration of our $\mathcal{D}_{\text{M-DPO}}$ construction process is in Fig~\ref{fig:main_fig}.~(c).

\paragraph{Training objective of M-DPO.}
With $\mathcal{D}_{\text{M-DPO}}$, the VideoLLM is further trained to align with the metric preferences.
Formally, the M-DPO loss regarding a single sample composed of ($\hat{y}^w_k$, $\hat{y}^l_k$, $m_k^w$, $m_k^l$, $v$, $h_{< k}$, $x_k$) is defined as:
\begin{equation}
\begin{split}
    & \mathcal{L}_{s}(\hat{y}^w_{k}, \hat{y}^l_{k}; v, h_{<k},x_k) = \\
    & \left[\log \sigma \left(\beta  r(\hat{y}^{w}_{k}; v, h_{< k},x_k) - \beta r(\hat{y}^{l}_{k}; v,h_{< k},x_k) \right) \right],
    \label{eq:sample-loss-M-DPO}
\end{split}
\end{equation}
where $\sigma$ denotes the sigmoid function, $r(\hat{y}_{k}; v, h_{< k},x_k) =  \log \frac{\pi_{\theta}\left( \hat{y}_{k}| v, h_{< k},x_k \right)}{\pi_{\text{ref}}\left(\hat{y}_{k}| v, h_{< k},x_k \right)}$ denotes a likelihood ratio, $\pi_{\theta}$ denotes the target model to be optimized, and $\beta$ is a hyperparameter controlling the distribution disparity of $\pi_{\theta}$ from the reference model $\pi_{\text{ref}}$.
Thus, by minimizing the given loss, it encourages the model to learn the metric-based preference on the $k$-th task by enlarging the gap of the likelihood ratio between preferred and dispreferred responses in terms of the target metric.
Then, the basic version of M-DPO training objective, $\mathcal{L}_{\text{M-DPO}^{-}}$, excluding the preference gap-aware module described in the following section, is defined as below, with the loss averaged across all samples in $\mathcal{D}_{\text{M-DPO}}$:
\begin{equation}
\begin{split}
    \mathcal{L}_{\text{M-DPO}^{-}} & (\pi_{\theta};  \pi_{\text{ref}}) = \\
    &- \mathbb{E}_{\mathcal{D}_{\text{M-DPO}}}
    \left[ \mathcal{L}_{s}\left(\hat{y}^w_{k}, \hat{y}^l_{k}; v,h_{<k},x_k\right)\right].
\end{split}
\label{eq:overall_dpo_loss}
\end{equation}
Note $\pi_{\theta}$ is built by adding LoRA modules after the initialization with $\pi_{\text{ref}}$, and LoRA modules in $\pi_{\theta}$ are only trainable parameters, and $\pi_{\text{ref}}$ is left unchanged.

\paragraph{Preference gap-aware M-DPO.}
Training data for conventional DPO only includes \textit{binary} preferences $\hat{y}^w$ and $\hat{y}^l$, which only indicates whether a response is preferred or not.
On the contrary, data in $\mathcal{D}_{\text{M-DPO}}$ also comprises \textit{continuous} preferences $m^w_{k}$ and $m^l_{k}$ which not only indicates whether a response is preferred or not but also reveals \textit{how much} as it is built on continuous metrics.
We observe that when optimizing with such continuous preferences, taking the gap of preferences between $\hat{y}^w_{k}$ and $\hat{y}^l_{k}$ into account further facilitates the proper training.
To this end, we propose $\mathcal{L}_{\text{M-DPO}}$, an advanced version of $\mathcal{L}_{\text{M-DPO}^{-}}$ by modifying Eq.~\eqref{eq:overall_dpo_loss} as:
\begin{equation}
\begin{split}
    & \mathcal{L}_{\text{M-DPO}} (\pi_{\theta};  \pi_{\text{ref}}) = - \mathbb{E}_{\mathcal{D}_{\text{M-DPO}}}  \\ 
    & \left[\mathds{1}\left(m^w_{k} - m^l_{k} > \gamma\right) \cdot \mathcal{L}_{s}\left(\hat{y}^w_{k}, \hat{y}^l_{k}; v, h_{<k},x_k\right)\right],
\end{split}
\label{eq:final_dpo_loss}
\end{equation}
where $\mathds{1}(\cdot)$ denotes an indicator function.
Concretely, we only calculate losses on preference pairs where the gap of the evaluation metrics between the preferred and dispreferred response is above a certain threshold $\gamma$, which is a hyperparameter.
Such an approach alleviates difficulties in optimizing pairs with subtle differences in metrics, thereby facilitating the overall optimization process.
In the following sections, the term \q{M-DPO} refers to $\mathcal{L}_{\text{M-DPO}}$ in Eq.~\eqref{eq:final_dpo_loss} instead of $\mathcal{L}_{\text{M-DPO}^{-}}$ in Eq.~\eqref{eq:overall_dpo_loss} unless specified.
Overall, we propose a novel framework named VidChain comprised of CoTasks and M-DPO which effectively enhances the fine-grained temporal video understanding of VideoLLMs.

\begin{table}[t!]
    \centering
    \small
    \renewcommand{\arraystretch}{1.1}
    \begin{adjustbox}{width=1.0\columnwidth}
    \begin{tabular}{l|cc|cc}
    \toprule  
      & \multicolumn{2}{c|}{ActivityNet} & \multicolumn{2}{c}{YouCook2}   \\ 
    & $\text{SODA}_c$ & METEOR & $\text{SODA}_c$ & METEOR  \\
    \hline
    \hline
    \rowcolor{lightgray}
    \multicolumn{5}{l}{\textbf{VTimeLLM}} \\ 
    \hline
    Baseline
        & 5.8  &  6.8 
        & 3.4 &  3.5    \\
        
    + CoTasks-$\mathcal{P}_{t \rightarrow c}$
        & 6.7  &  7.6
        & 4.1 &  4.4   \\
        
    + VidChain-$\mathcal{P}_{t \rightarrow c}$ 
       & \textbf{7.2} &  \textbf{7.7}  
        & \textbf{4.6} &  \textbf{4.9} \\
        
    \hline
    + CoTasks-$\mathcal{P}_{c \rightarrow t}$
        & 6.5  &  \textbf{7.4} 
        & 3.8 &  4.3    \\
        
    +  VidChain-$\mathcal{P}_{c \rightarrow t}$
        & \textbf{6.9} &  7.1  
        & \textbf{4.3} &  \textbf{4.5}   \\
        
    \hline
    \hline
    \rowcolor{lightgray}
    \multicolumn{5}{l}{\textbf{VideoLLaMA2}} \\
    \hline
    Baseline
        & 7.2 &  7.7   
        & 3.3 &  3.5    \\

    + CoTasks-$\mathcal{P}_{t \rightarrow c}$   
        & 7.5 &  8.3  
        & 4.2 &  5.1 \\
        
    +  VidChain-$\mathcal{P}_{t \rightarrow c}$  
        & \textbf{8.2} &  \textbf{8.7}
        & \textbf{4.6} &  \textbf{5.5}    \\
        
    \hline    
    + CoTasks-$\mathcal{P}_{c \rightarrow t}$
        & 7.7 &  8.5
        & 4.5 &  5.5  \\
        
    +  VidChain-$\mathcal{P}_{c \rightarrow t}$
        & \textbf{8.8} &  \textbf{8.8}    
        & \textbf{4.8} &  \textbf{5.6} \\

    \hline
    \end{tabular}
    \end{adjustbox}
    \caption{\textbf{Ablation study on components of VidChain.}
    VidChain denotes CoTasks + M-DPO. \newline
    }
    \vspace{-0.4cm}
    \label{tab:ablation}
\end{table}

\section{Experiments}

\begin{table}[t!]
\centering
\small
\begin{tabular}{@{}lcccc@{}}
\toprule
{Dataset} & {var.} & {Method} & {Acc($\uparrow$)} & {KL-Div($\downarrow$)} \\ 
\midrule
\multirow{2}{*}{ActivityNet} & \multirow{2}{*}{1.17} & Baseline & 40\% & 0.09 \\ 
 & & + VidChain & \textbf{70\%} & \textbf{0.04} \\ 
\midrule
\multirow{2}{*}{YouCook2} & \multirow{2}{*}{7.79} & Baseline & 8\% & 0.38 \\ 
 & & + VidChain & \textbf{16\%} & \textbf{0.06} \\ 
\bottomrule
\end{tabular}
\caption{\textbf{Performance on $p(n|v)$ of CoTasks}.
We evaluate segmentation accuracy and KL divergence with the ground truth distribution for datasets with distinct variances of the number of segments (denoted as var.).
}
\label{tab:subtask-seg}
\end{table}

\begin{table}[t!]
\centering
\small
\begin{tabular}{@{}lccc||ccc@{}}
\toprule
 & \multicolumn{3}{c||}{\textbf{$p(c|v,n)$}}  & \multicolumn{2}{c}{\textbf{$p(t|v,n)$}} \\ \midrule
 & {C} & {M} & {B} & {R@0.3} & {R@0.5} \\ 
\midrule
{Baseline} & 37.2 & 11.7 & 5.6 & 69.8 & 46.4\\ 
+ VidChain & \textbf{49.2} & \textbf{19.4} & \textbf{9.1} & \textbf{82.7} & \textbf{64.2}\\ 
\bottomrule
\end{tabular}
\caption{\textbf{Performance on $p(c|v,n)$ and $p(t|v,n)$ of CoTasks for $\mathcal{P}_{c\rightarrow t}$ and $\mathcal{P}_{t\rightarrow c}$ respectively.}
Note that C, M, and B denote CIDEr, METEOR, and BLEU4, respectively, and R@k denotes Recall@k.}
\label{tab:subtask-cap-and-time}
\end{table}

\subsection{Benchmarks}
\paragraph{Dense Video Captioning.}
We experiment on two different dense video captioning benchmarks, ActivityNet Captions~\cite{krishna2017dense} and YouCook2~\cite{zhou2018towards}.
As evaluation metrics, we adopt $\text{SODA}_c$~\cite{fujita2020soda}, METEOR~\cite{banerjee2005meteor}, and CIDEr~\cite{vedantam2015cider} following previous works~\cite{huang2024vtimellm,qian2024momentor}.

\paragraph{Temporal Video Grounding.}
For temporal video grounding, we use ActivityNet Captions dataset~\cite{krishna2017dense} to validate the effectiveness of our method.
For implementation details and further details about the datasets and their metrics, refer to the supplement.

\subsection{Main Results}
We evaluate VidChain on the challenging Dense Video Captioning (DVC) and temporal video grounding (TVG) to verify the effectiveness of our approach in enhancing fine-grained video understanding.
Note we report performances for both CoTasks paths, $\mathcal{P}_{c \rightarrow t}$ and $\mathcal{P}_{t \rightarrow c}$ for the DVC task.

\paragraph{Results.}
Tab.~\ref{tab:main} demonstrates the effectiveness of the proposed VidChain by applying it on two state-of-the-art VideoLLMs, VTimeLLM and VideoLLaMA2.
VidChain improves both VideoLLMs on two DVC benchmarks, ActivityNet and YouCook, thereby outperforming every VideoLLM.
In detail, VideoLLaMA2+VidChain-$\mathcal{P}_{c \rightarrow t}$ shows a 22.2\% gain in $\text{SODA}_c$ improving from 7.2 to 8.8, 14.3\% gain in METEOR and 33.4\% in CIDEr on ActivityNet.
In YouCook2, the model shows an 45.5\%, 60\%, 93.5\% increase for $\text{SODA}_c$, METEOR, and CIDEr, respectively.

Similar to the case of VideoLLaMA2, VidChain also shows consistent performance gains with VTimeLLM.
For instance, VidChain boosts performance of VTimeLLM by up to 1.2, 1.4, and 6.9 points in $\text{SODA}_c$, METEOR, and CIDEr respectively on the YouCook benchmark for DVC, while it also outperforms the baseline in the ActivityNet in every metrics.
Notably, VidChain applied to VTimeLLM with 7B LLM outperforms the baseline VTimeLLM with 13B LLM on every task by a large margin.

Moreover, Tab.~\ref{tab:main-tvg} demonstrates the effectiveness of VidChain on TVG, where we show a prominent increase in performance when applied to both VideoLLMs.
In particular, VTimeLLM+VidChain shows a 17.4, 16.0, 11.4, and 13.1 increase in Recall@0.3, Recall@0.5, Recall@0.7, and mIoU.
In addition, we outperform the model further trained with the corresponding TVG datasets.
The enhanced performance on TVG underlines the effectiveness of VidChain in enhancing the capability of a VideoLLM fine-grained video understanding, thereby also improving performance on a sub-task for DVC.
\begin{table}[t!]
    \centering
    \small
    \begin{adjustbox}{width=\linewidth}
    \renewcommand{\arraystretch}{1.0}
    \begin{tabular}{c|cc|cc}
        \toprule
        \multirow{2}{*}{} & \multicolumn{2}{c|}{Training Data ($\mathcal{D}_{\text{CT}}$)} & \multicolumn{2}{c}{Dense Video Captioning} \\
        & \large$\mathcal{D}_{t \rightarrow c}$ & \large$\mathcal{D}_{c \rightarrow t}$ & $\text{SODA}_c$ & METEOR  \\
        \midrule
        \midrule
        Baseline & \textcolor{BrickRed}{\ding{56}} & \textcolor{BrickRed}{\ding{56}} 
        & 7.2 & 7.7\\
        
        \midrule
       \multirow{2}{*}{
       CoTasks-\large$\mathcal{P}_{t \rightarrow c}$} & \textcolor{RoyalBlue}{\ding{52}} & \textcolor{BrickRed}{\ding{56}} 
       & 7.4 & 7.6  \\
       
        & \textcolor{RoyalBlue}{\ding{52}} & \textcolor{RoyalBlue}{\ding{52}} 
        & \textbf{7.5} & \textbf{8.3} \\
        \midrule
        
        \multirow{2}{*}{
        CoTasks-\large$\mathcal{P}_{c \rightarrow t}$} & \textcolor{BrickRed}{\ding{56}} & \textcolor{RoyalBlue}{\ding{52}} 
        & 7.6 & 8.1 \\
        
        & \textcolor{RoyalBlue}{\ding{52}} & \textcolor{RoyalBlue}{\ding{52}} & 
        \textbf{7.7} & \textbf{8.5} \\
        
        \bottomrule
    \end{tabular}
    \end{adjustbox}
    \caption{
    \textbf{Ablation study on data composition of $\mathcal{D}_{\text{CT}}$.}}
    \label{tab:cotasks_data_table}
\end{table}
\begin{table}[t!]
    \centering
    \small
    \setlength{\tabcolsep}{2mm}
    \begin{tabular}{l|cc|cc}
        \toprule
        &\multicolumn{2}{c|}{DVC}&\multicolumn{2}{c}{TVG}\\
         & $\text{SODA}_c$ & METEOR & R@0.3 & mIoU \\
        \midrule
        \midrule
         Baseline & 7.7 & 8.5 & 60.2 & 41.9 \\
        $\mathcal{L}_{\text{DPO}}$ & 8.3 & 8.6  & 61.6 & 42.8 \\
        $\mathcal{L}_{\text{M-DPO}^{-}}$ & 8.6 & \textbf{8.8}& 62.4 & 43.4 \\
        \rowcolor{lightblue}
        $\mathcal{L}_{\text{M-DPO}}$ \small(ours) & \textbf{8.8} & \textbf{8.8} & \textbf{63.3} & \textbf{44.1} \\
        \bottomrule
    \end{tabular}
    \caption{
    \textbf{Analysis on DPO objectives.}
    Note the baseline refers to VideoLLaMA2+CoTasks-$\mathcal{P}_{c \rightarrow t}$.
    }
    \label{tab:CMPO}
\end{table}

\begin{figure*}[t!]
    \centering
    \includegraphics[width=\textwidth]{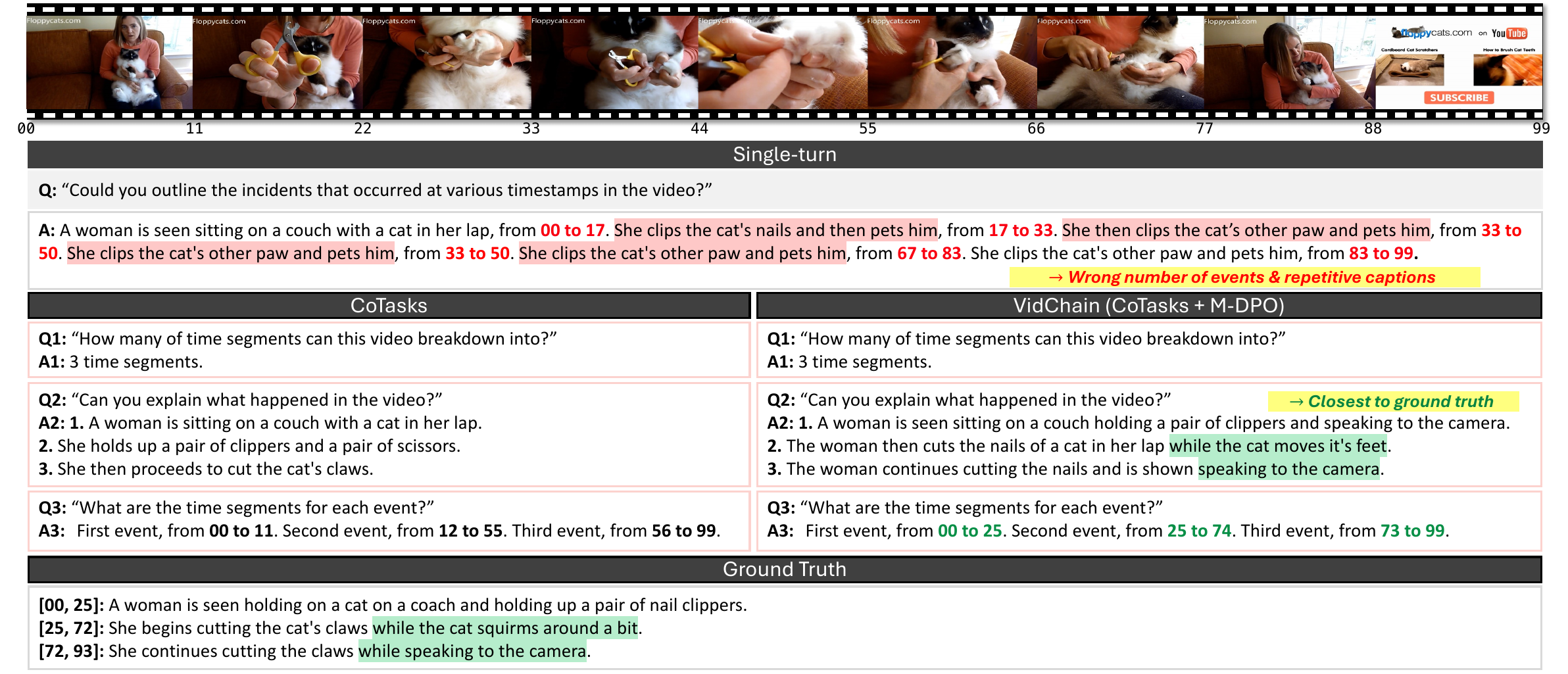}
    \caption{\textbf{Qualitative example of Dense Video Captioning.} 
    Predictions of baseline VideoLLM (Single-turn), VideoLLM+CoTasks, and VidChain (CoTasks + M-DPO) are illustrated.
    {Red} and {green} highlights denote erroneous and accurate predictions, respectively.
    Visualization is done on ActivityNet validation set with VTimeLLM in $\mathcal{P}_{c \rightarrow t}$ path.
    } 
    \label{fig:qual_vidchain}
\end{figure*}
\subsection{Quantitative Analysis}
In the following experiments and analysis, we report results on ActivityNet for DVC using our best-performing model VideoLLaMA2+VidChain unless specified.

\paragraph{Effectiveness of CoTasks.}
In Tab.~\ref{tab:ablation}, we analyze the effectiveness of CoTasks. 
Results show that CoTasks yields consistent performance improvement on both VTimeLLM and VideoLLaMA2 regardless of the inference path ($\mathcal{P}_{t \rightarrow c}$ or $\mathcal{P}_{c \rightarrow t}$), compared to the baselines. 
In particular, when applied to VTimeLLM, CoTasks-$\mathcal{P}_{t \rightarrow c}$ shows 15.5\% gain in $\text{SODA}_c$ for ActivityNet, and 20.6\% gain in YouCook2.
Moreover, we also observe consistent gains in TVG tasks by applying CoTasks, where the results are in the supplement.
This result verifies the effectiveness of CoTasks in enhancing the reasoning capability of a VideoLLM.

\paragraph{Effectiveness of M-DPO.}
In Tab.~\ref{tab:ablation}, results of +VidChain are also reported, which denotes that M-DPO is also applied on top of CoTasks.
Our results show that M-DPO generally improves performance on both VideoLLMs, VTimeLLM, and VideoLLaMA2.
For instance, further training VideoLLaMA2+CoTasks-$\mathcal{P}_{c \rightarrow t}$ with M-DPO improves $\text{SODA}_c$, and METEOR by 1.1, and 0.3 points, as shown by the result of VideoLLaMA2+VidChain-$\mathcal{P}_{c \rightarrow t}$.
Similarly, applying M-DPO consistently boosts performance on every VideoLLMs and benchmarks, showing its effectiveness.

\paragraph{Ablation studies on subtasks of CoTasks.}
\label{sup:subtasks-of-vidchain}
We evaluate performances on three subtasks that compose CoTasks for a detailed investigation: (1) time segment prediction $p(n|v)$, which is illustrated in the grey box in Fig.~\ref{fig:main_fig}, (2) video paragraph captioning $p(c|v, n)$ of $\mathcal{P}_{c \rightarrow t}$, illustrated as the first red box and (3) temporal video segmentation $p(t|v, n)$ of $\mathcal{P}_{t \rightarrow c}$, illustrated as the first yellow box. 
As shown in both Tab.~\ref{tab:subtask-seg} and Tab.~\ref{tab:subtask-cap-and-time}, among all subtasks, VidChain consistently achieves superior performance.
Note for (1) we conduct experiment on two benchmark datasets with different variances of the number of segments in the video.
For (3), we conducted experiment with the samples where both the baseline and VidChain accurately predicted the number of segments.

\paragraph{Ablation study on data composition of $\mathcal{D}_{\text{CT}}$.}
We conduct an ablation study on the effect of the inclusion of data with two paths for DVC, namely $\mathcal{D}_{t \rightarrow c}$ and $\mathcal{D}_{c \rightarrow t}$ in CoTasks training data $\mathcal{D}_{\text{CT}}$.
The results are in Tab.~\ref{tab:cotasks_data_table}, where including data with both paths is shown to perform better than using only a single type of path.
We conjecture that two different paths are complementary to each other, therefore composing $\mathcal{D}_{\text{CT}}$ with data in both paths facilitates a VideoLLM's fine-grained video understanding capacity by letting a VideoLLM learn to solve the same objective in different ways.

\paragraph{Analysis on DPO objectives.}
In Tab.~\ref{tab:CMPO}, we analyze different DPO objectives using our $\mathcal{D}_{\text{M-DPO}}$ dataset, where the optimization objective applied is only different.
$\mathcal{L}_{\text{DPO}}$ (row 2) denotes that DPO is only applied to the final task instead of intermediate tasks and preference-gap aware DPO is not applied.
Overall, it generally improves DVC performance, showing the effectiveness of the DPO approach in the DVC task.
Still, additionally optimizing intermediate tasks with $\mathcal{L}_{\text{M-DPO}^{-}}$ (row 3) enables additional gains of 0.3, and 0.2 over $\mathcal{L}_{\text{DPO}}$~(row 2) in $\text{SODA}_c$, and METEOR.
Finally, applying the preference gap-aware M-DPO ($\mathcal{L}_{\text{M-DPO}}$) results in the best performance (row 4).
The ablation results show the effectiveness of M-DPO components.
A more detailed explanation, \textit{e.g.}, DPO margin for each method across training iteration, is in the supplement.

\subsection{Qualitative Analysis}
Fig.~\ref{fig:qual_vidchain} compares qualitative results of a baseline VideoLLM (single-turn) with one trained using CoTasks and VidChain (CoTasks + M-DPO).
As illustrated, baseline VideoLLM shows inferior performance on DVC, segmenting a video into an overly large number of events (6 predicted vs. 3 ground-truth), while captions for each event are also highly repetitive (\dq{She clips the cat's other paw and pets him}), revealing the lack of a baseline in the capability of fine-grained video understanding.
In contrast, a VideoLLM trained with CoTasks successfully segments a video into three events, while captions for each segment are more distinctive, showing the effectiveness in breaking down complex DVC into sub-tasks.
Moreover, further aligning the model with M-DPO yields the best results, closely matching ground-truths and showcasing M-DPO's metric-aligned supervision.

\section{Conclusion}
In this paper, we propose a framework, VidChain, comprised of the Chain-of-Tasks (CoTasks), and Metric-based Direct Preference Optimization (M-DPO).
CoTasks decompose complicated tasks into series of sub-tasks, easing the difficulties of solving the task.
M-DPO aligns a VideoLLM with evaluation metrics of sub-tasks, providing supervision aligned with the abilities required for those tasks.
Applied on two different VideoLLMs, VidChain enhances fine-grained understanding of models, consistently improving their performance thereby outperforming previous VideoLLMs.

\section*{Acknowledgments}
This work was partly supported by the National Supercomputing Center with supercomputing resources including technical support (KSC-2023-CRE-0185, 25\%), and Institute of Information \& Communications Technology Planning \& Evaluation (IITP) grants funded by the Korea government (MSIP \& MSIT) (No. RS-2024-00443251, 25\%, Accurate and Safe Multimodal, Multilingual Personalized AI Tutors; No. RS-2024-00457882, 25\%, National AI Research Lab Project; No. IITP-2024-RS-2024-00436857, 25\%, Information Technology Research Center ITRC).

\bibliography{aaai25}

\begin{thebibliography}{38}
\providecommand{\natexlab}[1]{#1}

\bibitem[{Ahn et~al.(2024)Ahn, Choi, Yu, Kang, and Choi}]{ahn2024tuning}
Ahn, D.; Choi, Y.; Yu, Y.; Kang, D.; and Choi, J. 2024.
\newblock Tuning large multimodal models for videos using reinforcement learning from ai feedback.
\newblock In \emph{ACL}.

\bibitem[{Banerjee and Lavie(2005)}]{banerjee2005meteor}
Banerjee, S.; and Lavie, A. 2005.
\newblock METEOR: An automatic metric for MT evaluation with improved correlation with human judgments.
\newblock In \emph{ACL}.

\bibitem[{Chen et~al.(2024)Chen, Wu, Wang, Su, Chen, Xing, Zhong, Zhang, Zhu, Lu et~al.}]{chen2024internvl}
Chen, Z.; Wu, J.; Wang, W.; Su, W.; Chen, G.; Xing, S.; Zhong, M.; Zhang, Q.; Zhu, X.; Lu, L.; et~al. 2024.
\newblock Internvl: Scaling up vision foundation models and aligning for generic visual-linguistic tasks.
\newblock In \emph{CVPR}.

\bibitem[{Cheng et~al.(2024)Cheng, Leng, Zhang, Xin, Li, Chen, Zhu, Zhang, Luo, Zhao et~al.}]{cheng2024videollama}
Cheng, Z.; Leng, S.; Zhang, H.; Xin, Y.; Li, X.; Chen, G.; Zhu, Y.; Zhang, W.; Luo, Z.; Zhao, D.; et~al. 2024.
\newblock VideoLLaMA 2: Advancing Spatial-Temporal Modeling and Audio Understanding in Video-LLMs.
\newblock arXiv:2406.07476.

\bibitem[{Christiano et~al.(2017)Christiano, Leike, Brown, Martic, Legg, and Amodei}]{christiano2017deep}
Christiano, P.~F.; Leike, J.; Brown, T.; Martic, M.; Legg, S.; and Amodei, D. 2017.
\newblock Deep reinforcement learning from human preferences.
\newblock In \emph{NeurIPS}.

\bibitem[{Dai et~al.(2023)Dai, Li, Li, Tiong, Zhao, Wang, Li, Fung, and Hoi}]{dai2023instructblip}
Dai, W.; Li, J.; Li, D.; Tiong, A.; Zhao, J.; Wang, W.; Li, B.; Fung, P.; and Hoi, S. 2023.
\newblock Instructblip: Towards general-purpose vision-language models with instruction tuning.
\newblock In \emph{NeurIPS}.

\bibitem[{Fujita et~al.(2020)Fujita, Hirao, Kamigaito, Okumura, and Nagata}]{fujita2020soda}
Fujita, S.; Hirao, T.; Kamigaito, H.; Okumura, M.; and Nagata, M. 2020.
\newblock SODA: Story oriented dense video captioning evaluation framework.
\newblock In \emph{ECCV}.

\bibitem[{Gunjal, Yin, and Bas(2024)}]{gunjal2024detecting}
Gunjal, A.; Yin, J.; and Bas, E. 2024.
\newblock Detecting and preventing hallucinations in large vision language models.
\newblock In \emph{AAAI}.

\bibitem[{Huang et~al.(2024)Huang, Wang, Chen, Song, and Zhu}]{huang2024vtimellm}
Huang, B.; Wang, X.; Chen, H.; Song, Z.; and Zhu, W. 2024.
\newblock Vtimellm: Empower llm to grasp video moments.
\newblock In \emph{CVPR}.

\bibitem[{Kim et~al.(2024)Kim, Kim, Moon, Choi, and Kim}]{kim2024you}
Kim, M.; Kim, H.~B.; Moon, J.; Choi, J.; and Kim, S.~T. 2024.
\newblock Do You Remember? Dense Video Captioning with Cross-Modal Memory Retrieval.
\newblock In \emph{CVPR}.

\bibitem[{Ko et~al.(2023{\natexlab{a}})Ko, Lee, Choi, Chu, Park, and Kim}]{ko2023open}
Ko, D.; Lee, J.~S.; Choi, M.; Chu, J.; Park, J.; and Kim, H.~J. 2023{\natexlab{a}}.
\newblock Open-Vocabulary Video Question Answering: A New Benchmark for Evaluating the Generalizability of Video Question Answering Models.
\newblock In \emph{ICCV}.

\bibitem[{Ko et~al.(2023{\natexlab{b}})Ko, Lee, Kang, Roh, and Kim}]{ko2023large}
Ko, D.; Lee, J.~S.; Kang, W.; Roh, B.; and Kim, H.~J. 2023{\natexlab{b}}.
\newblock Large language models are temporal and causal reasoners for video question answering.
\newblock In \emph{EMNLP}.

\bibitem[{Krishna et~al.(2017)Krishna, Hata, Ren, Fei-Fei, and Carlos~Niebles}]{krishna2017dense}
Krishna, R.; Hata, K.; Ren, F.; Fei-Fei, L.; and Carlos~Niebles, J. 2017.
\newblock Dense-captioning events in videos.
\newblock In \emph{ICCV}.

\bibitem[{Lai et~al.(2024)Lai, Tian, Chen, Yang, Peng, and Jia}]{lai2024step}
Lai, X.; Tian, Z.; Chen, Y.; Yang, S.; Peng, X.; and Jia, J. 2024.
\newblock Step-DPO: Step-wise Preference Optimization for Long-chain Reasoning of LLMs.
\newblock arXiv:2406.18629.

\bibitem[{Li et~al.(2023)Li, He, Wang, Li, Wang, Luo, Wang, Wang, and Qiao}]{li2023videochat}
Li, K.; He, Y.; Wang, Y.; Li, Y.; Wang, W.; Luo, P.; Wang, Y.; Wang, L.; and Qiao, Y. 2023.
\newblock Videochat: Chat-centric video understanding.
\newblock arXiv:2305.06355.

\bibitem[{Li et~al.(2024)Li, Wang, He, Li, Wang, Liu, Wang, Xu, Chen, Luo et~al.}]{li2024mvbench}
Li, K.; Wang, Y.; He, Y.; Li, Y.; Wang, Y.; Liu, Y.; Wang, Z.; Xu, J.; Chen, G.; Luo, P.; et~al. 2024.
\newblock Mvbench: A comprehensive multi-modal video understanding benchmark.
\newblock In \emph{CVPR}.

\bibitem[{Lin et~al.(2023)Lin, Zhu, Ye, Ning, Jin, and Yuan}]{lin2023video}
Lin, B.; Zhu, B.; Ye, Y.; Ning, M.; Jin, P.; and Yuan, L. 2023.
\newblock Video-llava: Learning united visual representation by alignment before projection.

\bibitem[{Liu et~al.(2024)Liu, Li, Li, and Lee}]{liu2024improved}
Liu, H.; Li, C.; Li, Y.; and Lee, Y.~J. 2024.
\newblock Improved baselines with visual instruction tuning.
\newblock In \emph{CVPR}.

\bibitem[{Liu et~al.(2023)Liu, Li, Wu, and Lee}]{liu2023visual}
Liu, H.; Li, C.; Wu, Q.; and Lee, Y.~J. 2023.
\newblock Visual instruction tuning.
\newblock In \emph{NeurIPS}.

\bibitem[{Maaz et~al.(2024)Maaz, Rasheed, Khan, and Khan}]{maaz2023video}
Maaz, M.; Rasheed, H.; Khan, S.; and Khan, F.~S. 2024.
\newblock Video-chatgpt: Towards detailed video understanding via large vision and language models.
\newblock In \emph{ACL}.

\bibitem[{Ouyang et~al.(2022)Ouyang, Wu, Jiang, Almeida, Wainwright, Mishkin, Zhang, Agarwal, Slama, Ray et~al.}]{ouyang2022training}
Ouyang, L.; Wu, J.; Jiang, X.; Almeida, D.; Wainwright, C.; Mishkin, P.; Zhang, C.; Agarwal, S.; Slama, K.; Ray, A.; et~al. 2022.
\newblock Training language models to follow instructions with human feedback.
\newblock In \emph{NeurIPS}.

\bibitem[{Qian et~al.(2024)Qian, Li, Wu, Ye, Fei, Chua, Zhuang, and Tang}]{qian2024momentor}
Qian, L.; Li, J.; Wu, Y.; Ye, Y.; Fei, H.; Chua, T.-S.; Zhuang, Y.; and Tang, S. 2024.
\newblock Momentor: Advancing video large language model with fine-grained temporal reasoning.
\newblock In \emph{ICML}.

\bibitem[{Rafailov et~al.(2023)Rafailov, Sharma, Mitchell, Manning, Ermon, and Finn}]{rafailov2024direct}
Rafailov, R.; Sharma, A.; Mitchell, E.; Manning, C.~D.; Ermon, S.; and Finn, C. 2023.
\newblock Direct preference optimization: Your language model is secretly a reward model.
\newblock In \emph{NeurIPS}.

\bibitem[{Ren et~al.(2024)Ren, Yao, Li, Sun, and Hou}]{ren2024timechat}
Ren, S.; Yao, L.; Li, S.; Sun, X.; and Hou, L. 2024.
\newblock Timechat: A time-sensitive multimodal large language model for long video understanding.
\newblock In \emph{CVPR}.

\bibitem[{Song et~al.(2024)Song, Yu, Li, Yu, Huang, Li, and Wang}]{song2024preference}
Song, F.; Yu, B.; Li, M.; Yu, H.; Huang, F.; Li, Y.; and Wang, H. 2024.
\newblock Preference ranking optimization for human alignment.
\newblock In \emph{AAAI}.

\bibitem[{Vedantam, Lawrence~Zitnick, and Parikh(2015)}]{vedantam2015cider}
Vedantam, R.; Lawrence~Zitnick, C.; and Parikh, D. 2015.
\newblock Cider: Consensus-based image description evaluation.
\newblock In \emph{CVPR}.

\bibitem[{Wang et~al.(2021)Wang, Zhang, Lu, Zheng, Cheng, and Luo}]{wang2021end}
Wang, T.; Zhang, R.; Lu, Z.; Zheng, F.; Cheng, R.; and Luo, P. 2021.
\newblock End-to-end dense video captioning with parallel decoding.
\newblock In \emph{ICCV}.

\bibitem[{Xu et~al.(2024)Xu, Sharaf, Chen, Tan, Shen, Van~Durme, Murray, and Kim}]{xu2024contrastive}
Xu, H.; Sharaf, A.; Chen, Y.; Tan, W.; Shen, L.; Van~Durme, B.; Murray, K.; and Kim, Y.~J. 2024.
\newblock Contrastive preference optimization: Pushing the boundaries of llm performance in machine translation.
\newblock In \emph{ICML}.

\bibitem[{Yang et~al.(2024)Yang, Nagrani, Laptev, Sivic, and Schmid}]{yang2024vidchapters}
Yang, A.; Nagrani, A.; Laptev, I.; Sivic, J.; and Schmid, C. 2024.
\newblock Vidchapters-7m: Video chapters at scale.
\newblock In \emph{NeurIPS}.

\bibitem[{Yang et~al.(2023)Yang, Nagrani, Seo, Miech, Pont-Tuset, Laptev, Sivic, and Schmid}]{yang2023vid2seq}
Yang, A.; Nagrani, A.; Seo, P.~H.; Miech, A.; Pont-Tuset, J.; Laptev, I.; Sivic, J.; and Schmid, C. 2023.
\newblock Vid2seq: Large-scale pretraining of a visual language model for dense video captioning.
\newblock In \emph{CVPR}.

\bibitem[{Ye et~al.(2024)Ye, Xu, Ye, Yan, Hu, Liu, Qian, Zhang, and Huang}]{ye2024mplug}
Ye, Q.; Xu, H.; Ye, J.; Yan, M.; Hu, A.; Liu, H.; Qian, Q.; Zhang, J.; and Huang, F. 2024.
\newblock mplug-owl2: Revolutionizing multi-modal large language model with modality collaboration.
\newblock In \emph{CVPR}.

\bibitem[{Yu et~al.(2024)Yu, Yao, Zhang, He, Han, Cui, Hu, Liu, Zheng, Sun et~al.}]{yu2024rlhf}
Yu, T.; Yao, Y.; Zhang, H.; He, T.; Han, Y.; Cui, G.; Hu, J.; Liu, Z.; Zheng, H.-T.; Sun, M.; et~al. 2024.
\newblock Rlhf-v: Towards trustworthy mllms via behavior alignment from fine-grained correctional human feedback.
\newblock In \emph{CVPR}.

\bibitem[{Yuan et~al.(2024)Yuan, Cui, Wang, Ding, Wang, Deng, Shan, Chen, Xie, Lin et~al.}]{yuan2024advancing}
Yuan, L.; Cui, G.; Wang, H.; Ding, N.; Wang, X.; Deng, J.; Shan, B.; Chen, H.; Xie, R.; Lin, Y.; et~al. 2024.
\newblock Advancing llm reasoning generalists with preference trees.
\newblock arXiv:2404.02078.

\bibitem[{Zhang, Li, and Bing(2023)}]{zhang2023video}
Zhang, H.; Li, X.; and Bing, L. 2023.
\newblock Video-llama: An instruction-tuned audio-visual language model for video understanding.
\newblock In \emph{EMNLP-Demo}.

\bibitem[{Zhou, Xu, and Corso(2018)}]{zhou2018towards}
Zhou, L.; Xu, C.; and Corso, J. 2018.
\newblock Towards automatic learning of procedures from web instructional videos.
\newblock In \emph{AAAI}.

\bibitem[{Zhou et~al.(2024)Zhou, Arnab, Buch, Yan, Myers, Xiong, Nagrani, and Schmid}]{zhou2024streaming}
Zhou, X.; Arnab, A.; Buch, S.; Yan, S.; Myers, A.; Xiong, X.; Nagrani, A.; and Schmid, C. 2024.
\newblock Streaming dense video captioning.
\newblock In \emph{CVPR}.

\bibitem[{Zhu et~al.(2024)Zhu, Lin, Ning, Yan, Cui, Wang, Pang, Jiang, Zhang, Li et~al.}]{zhu2023languagebind}
Zhu, B.; Lin, B.; Ning, M.; Yan, Y.; Cui, J.; Wang, H.; Pang, Y.; Jiang, W.; Zhang, J.; Li, Z.; et~al. 2024.
\newblock LanguageBind: Extending Video-Language Pretraining to N-modality by Language-based Semantic Alignment.
\newblock In \emph{ICLR}.

\bibitem[{Zhu et~al.(2021)Zhu, Su, Lu, Li, Wang, and Dai}]{zhu2020deformable}
Zhu, X.; Su, W.; Lu, L.; Li, B.; Wang, X.; and Dai, J. 2021.
\newblock Deformable detr: Deformable transformers for end-to-end object detection.
\newblock In \emph{ICLR}.

\end{thebibliography}

\clearpage
\appendix
\section*{\Large{Appendix}}

The appendix is organized into the following sections:
\begin{itemize}
    \item Appendix~\ref{sup:cotasks_prompt_temp}: CoTasks prompt templates
    \item Appendix~\ref{sup:gamma_analysis}: Sensitivity analysis on $\gamma$
    \item Appendix~\ref{supp:dpo-margin}: Analysis on DPO margin.
    \item Appendix~\ref{sup:dpo_generate}: M-DPO data generated with predictions
    \item Appendix~\ref{sup:mdpo_vs_llm_loss}: M-DPO vs. LLM loss
    \item Appendix~\ref{sup:ablation_tvg}: Ablation on VidChain for TVG
    \item Appendix~\ref{sup:inference-cost}: Inference Cost
    \item Appendix~\ref{supp:benchmark-detail}: Benchmark Details
    \item Appendix~\ref{sup:CoTasks_dataset}: Details on $\mathcal{D}_{\text{CT}}$
    \item Appendix~\ref{sup:metric_detail}: Metric Details
    \item Appendix~\ref{sup:training_details}: Training Details
    \item Appendix~\ref{sup:qualitatives}: Further qualitative examples
\end{itemize}

\section{CoTasks prompt templates}
\label{sup:cotasks_prompt_temp}
\begin{figure}[h!]
    \centering
    \includegraphics[width=1.0\linewidth]{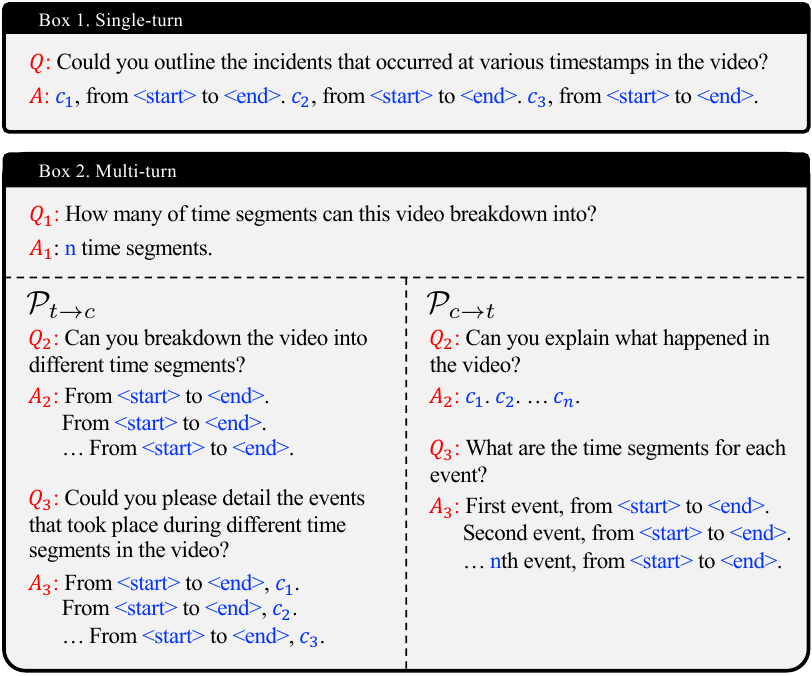}
    \caption{\textbf{CoTasks prompt template for DVC.}}
    \label{supp-fig:prompt}
\end{figure}
Figure~\ref{supp-fig:prompt} illustrates the prompts adopted in CoTasks for both reasoning paths of $\mathcal{P}_{t \rightarrow c}$ and $\mathcal{P}_{c \rightarrow t}$ and the single-turn prompt specifically for DVC.
For the single-turn (Box 1), we simply prompt the model to `outline the incidents that occurred at various timestamps', and the answer is given as an interleaved form of each event caption and start and end time boundaries. 

In the multi-turn, both reasoning paths initially start with prompting `How many of time segments can this video breakdown into?'.
Then for $\mathcal{P}_{t \rightarrow c}$, we prompt `Can you breakdown the video into different time segments?', where the answer is given to list the time boundaries for each event, \textit{e.g.}, `From $\langle$start$\rangle$ to $\langle$end$\rangle$. From $\langle$start$\rangle$ to $\langle$end$\rangle\ldots$', the last task prompt is to generate the captions for each event given the time boundaries, which also adopts the interleaved answer format similar to the single-turn.
In contrast, $\mathcal{P}_{c \rightarrow t}$ first prompts to generate the captions for each event first, `Can you explain what happened in the video?' and the answer is given as a list of captions.
Then the next task prompt is `What are the time segments for each event?', and instead of the interleaved form of the answer, we simply design to answer with a list of event time boundaries considering the sequence length of the model.
Note the start and end time boundaries ranges from 00 to 99, following VtimeLLM~\cite{huang2024vtimellm}.

\section{Sensitivity analysis on $\gamma$}
\label{sup:gamma_analysis}

\begin{table}[t!]
    \centering
    \begin{adjustbox}{width=0.9\linewidth}
    \small
    \begin{tabular}{c|cc|cc}
        \toprule
         &\multicolumn{2}{c|}{DVC}&\multicolumn{2}{c}{TVG}\\
        $\gamma$ & $\text{SODA}_c$ & METEOR & R@0.3 & mIoU \\

        \midrule
        \midrule
        5.0 & 8.8 & 8.4 & 61.1 & 42.6 \\
        7.5 & 8.9 & 8.5 & 61.1 & 42.7 \\
        10.0 & 8.8 & \textbf{8.8} & \textbf{63.3} & \textbf{44.1} \\
        12.5 & \textbf{9.0} & 8.7 & 61.2 & 43.0 \\
        15.0 & 8.8 & 8.5 & 61.2 & 42.9 \\
       
        \bottomrule
    \end{tabular}
    \end{adjustbox}
    \caption{\textbf{Sensitivity analysis on $\gamma$.} 
    We conduct experiments with VideoLLaMA2+VidChain-$\mathcal{P}_{c \rightarrow t}$.
    }
    \label{supp-tab:threshold_ablation_tab}
\end{table}
In Tab.~\ref{supp-tab:threshold_ablation_tab}, we show sensitivity analysis of M-DPO on $\gamma$ that is the threshold for the gap of the evaluation metrics between the preferred and dispreferred response in $\mathcal{L}_\text{M-DPO}$ of which is defined as follows:
\begin{equation}
\begin{split}
    & \mathcal{L}_{\text{M-DPO}} (\pi_{\theta};  \pi_{\text{ref}}) = - \mathbb{E}_{d_{\text{M-DPO}} \sim \mathcal{D}_{\text{M-DPO}}}  \\ 
    &\left[\mathds{1}\left(m^w_{k} - m^l_{k} > \gamma\right) \cdot \mathcal{L}_{s}\left(\hat{y}^w_{k}, \hat{y}^l_{k}; v, h_{<k},x_k\right)\right].
\end{split}
\label{eq:final_dpo_loss_supp}
\end{equation}
We conduct experiments ranging from $\gamma=5.0$ to $\gamma=15.0$, where the results show to be less sensitive to the change of $\gamma$.
We find that $\gamma$ = 10.0 (row 3) shows the best performance, resulting in slightly higher results in terms of METEOR for DVC, and R@0.3, mIoU for TVG.

Interestingly, we also observed when constructing the preference dataset, enlarging the sampling rate of hard negatives (\textit{e.g.}, $10 \le \gamma \le 20$) can result in further performance improvement.
This suggests that a higher proportion of challenging samples in training can better guide the model toward robust preference alignment.

\section{Analysis on DPO margin.}
\label{supp:dpo-margin}

\begin{figure}[t!]
    \centering
    \includegraphics[width=0.85\linewidth]{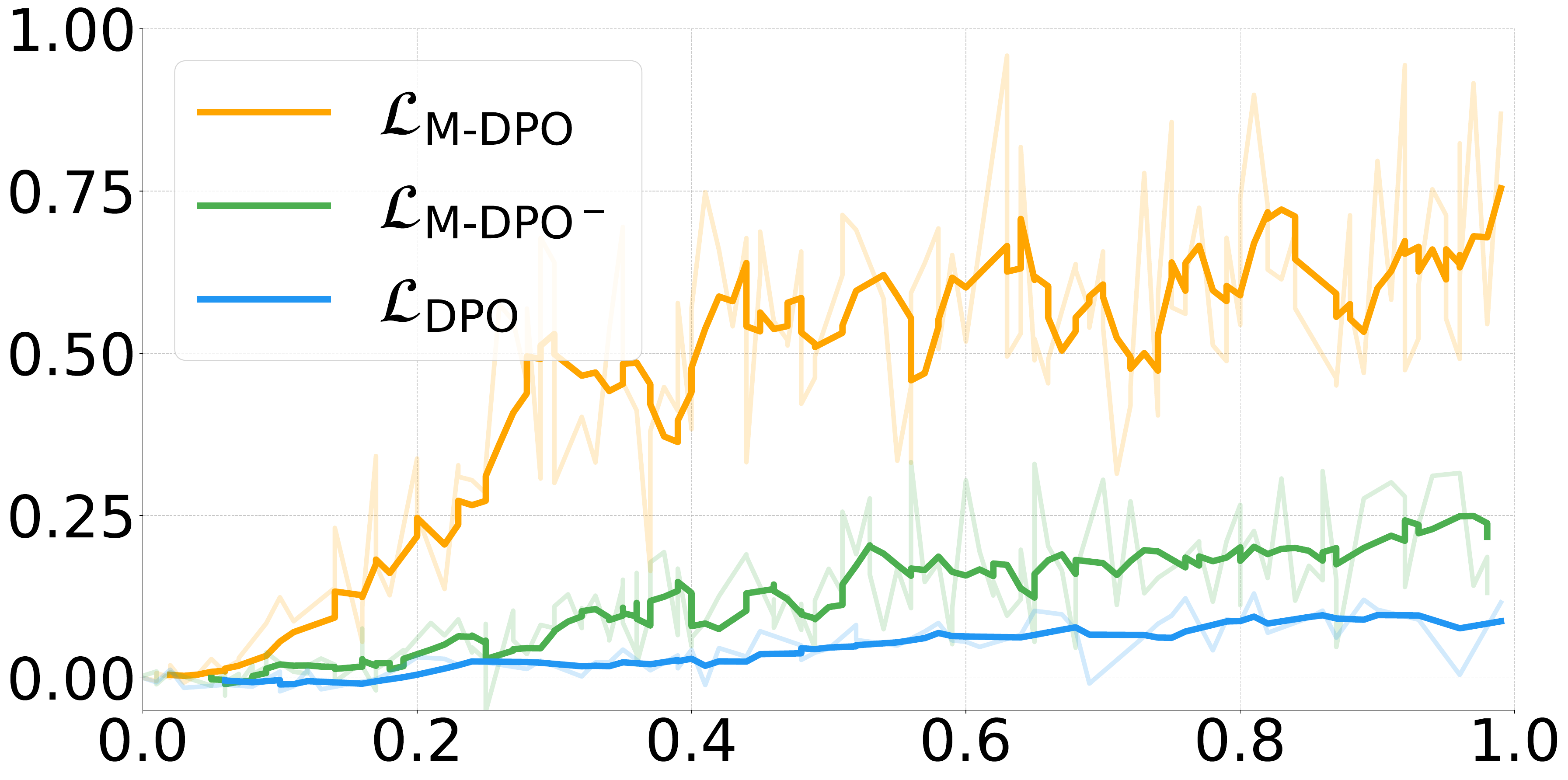}
    \caption{\textbf{Margin of the likelihood ratio between preferred and dispreferred responses} with $\mathcal{L}_{\text{DPO}}$, $\mathcal{L}_{\text{M-DPO}^{-}}$, and $\mathcal{L}_{\text{M-DPO}}$. $x$-axis stands for training epochs.}
    \label{fig:margin_fig}
\end{figure}

In Fig.~\ref{fig:margin_fig}, the margin of likelihood ratio between preferred and dispreferred responses under different DPO objectives is plotted by epoch.
A larger margin implies that preferred and dispreferred responses are clearly distinguished by a model.
As illustrated, $\mathcal{L}_{\text{DPO}}$ ({blue}) fails to teach the model to discriminate between preferred and dispreferred responses, as shown by the smallest margin.
On the contrary, also optimizing the intermediate task with $\mathcal{L}_{\text{M-DPO}^{-}}$ further enlarges the margin between responses ({green}).
Furthermore, only optimizing samples where the preference gap between responses is large enough with $\mathcal{L}_{\text{M-DPO}}$ ({orange}) results in the largest margin, which is 6.6 times that in $\mathcal{L}_{\text{DPO}}$ at the end of training.
The results show that each component in M-DPO contributes to teaching a VideoLLM to better discriminate between preferred and dispreferred responses.

\begin{table}[t!]
    \centering
    \begin{adjustbox}{width=0.80\linewidth}
    \small
    \begin{tabular}{l|cc}
        \toprule
        & \multicolumn{2}{c}{Dense Video Captioning } \\
          & $\text{SODA}_c$ & METEOR  \\
        \midrule
        \midrule

       CoTasks-$\mathcal{P}_{t \rightarrow c}$
       & 7.5 & 8.3 \\
        + $\mathcal{L}_{\text{LLM}}$
        & 7.4 & 8.6 \\

        \rowcolor{lightblue}
        + $\mathcal{L}_{\text{M-DPO}}$
        & \textbf{8.2} & \textbf{8.7} \\

        \midrule
        
        CoTasks-$\mathcal{P}_{c \rightarrow t}$ 
        & 7.7 & 8.5 \\
        + $\mathcal{L}_{\text{LLM}}$
        & 7.7 & 8.6 \\

        \rowcolor{lightblue}
         + $\mathcal{L}_{\text{M-DPO}}$
        & \textbf{8.8} & \textbf{8.8} \\

        \bottomrule
    \end{tabular}
    \end{adjustbox}
    \caption{
    \textbf{M-DPO vs LLM next-token prediction Loss.} 
    Note we adopt VideoLLaMA2 as the baseline model and evaluated on ActivityNet for DVC.}
    \label{supp-tab:llm-loss}
    \vspace{-0.2cm}
\end{table}

\section{M-DPO vs. LLM Loss}
\label{sup:mdpo_vs_llm_loss}
In Tab~\ref{supp-tab:llm-loss}, we compare DVC performance on fine-tuning the VideoLLM with $\mathcal{L}_{\text{M-DPO}}$ and with $\mathcal{L}_{\text{LLM}}$ that is the next-token prediction loss, \textit{i.e.}, cross-entropy loss.
Specifically, when finetuning with $\mathcal{L}_{\text{LLM}}$, we only use the preferred samples among the preferred and dispreferred paired samples from the $\mathcal{D}_{\text{M-DPO}}$.
The results show that simply finetuning with the next-token prediction loss is insufficient to cover the complex evaluation protocol that DVC requires.
In particular, VideoLLaMA2+CoTasks-$\mathcal{P}_{t \rightarrow c}$ shows 0.1 decrease and 0.3 increase in terms of $\text{SODA}_c$ and METEOR, whereas when fine-tuned with our $\mathcal{L}_{\text{M-DPO}}$ results in 0.8 and 0.1 increase in $\text{SODA}_c$ and METEOR compared to the baseline, respectively.
Hence, this validates the effectiveness of M-DPO in comparison to simple next-token prediction loss.


\begin{table}[t!]
    \centering
    \small
    \renewcommand{\arraystretch}{1.1}
    \begin{adjustbox}{width=0.90\columnwidth}
    \begin{tabular}{l|cccc}
    \toprule  
      & \multicolumn{4}{c}{Temporal Video Grounding (TVG)}  \\ 
    & R@0.3 & R@0.5 & R@0.7 & mIoU  \\
    \hline
    \hline
    \rowcolor{lightgray}
    \multicolumn{5}{l}{\textbf{VTimeLLM}} \\ 
    \hline
    Baseline
        & 44.0  &  27.8 
        & 14.3 &  30.4    \\
        
    + CoTasks
        & 58.6  &  41.0 
        & 23.8 &  41.2   \\
    + VidChain 
       & \textbf{61.4} &  \textbf{43.8}  
        & \textbf{25.7} &  \textbf{43.5} \\

    \hline
    $\text{Baseline}^{*}$~(TVG only) & 55.8 &  35.0  & 18.9  & 37.9 \\
        
    \hline
    \hline
    \rowcolor{lightgray}
    \multicolumn{5}{l}{\textbf{VideoLLaMA2}} \\
    \hline
    Baseline
        & 49.4 &  26.8   
        & 15.0 &  33.9    \\
        
    + CoTasks   
        & 60.2 &  41.5 
        & 23.2 &  42.0 \\
        
    +  VidChain  
        & \textbf{63.3} &  \textbf{44.8}
        & \textbf{25.2} &  \textbf{44.1}    \\
    \hline
    $\text{Baseline}^{*}$~(TVG only) & 59.9 &  41.5  & 22.5  & 42.2 \\
    \hline
    \end{tabular}
    \end{adjustbox}
    \vspace{-0.2cm}
    \caption{\textbf{Ablation study on components of VidChain for TVG.}
    VidChain denotes CoTasks + M-DPO. 
     * indicates the model further trained on the corresponding TVG dataset.
    }
    \label{supp-tab:ablation-tvg}
\end{table}
\section{Ablation study on VidChain for TVG}
\label{sup:ablation_tvg}
Tab.~\ref{supp-tab:ablation-tvg} shows the ablation study on components of VidChain for TVG and the ablation study on DVC is in our main paper Tab. 3.
Initially, finetuning with our CoTasks approach shows large performance gains for both VideoLLMs~\cite{huang2024vtimellm} of VTimeLLM and VideoLLaMA2~\cite{cheng2024videollama}.
For instance, VideoLLaMA2+CoTasks shows 10.8 points, 14.7 points, 8.2 points, and 8.1 points gains for R@0.3, R@0.5, R@0.7, and mIoU respectively, which is on par with the model further trained on the corresponding TVG dataset.
Furthermore, the adoption of M-DPO enhances the performance further, where VideoLLaMA2+VidChain shows 3.1, 3.3, 2, and 2.1 points gains respectively.
Notably, for M-DPO, task samples beyond DVC were not explicitly included. 
Nonetheless, the supervision from M-DPO on the sub-tasks of CoTasks led to notable performance improvements in TVG.
Overall, each component of VidChain is effective in enhancing the performance of the VideoLLM not only for DVC but also for TVG.

\begin{table}[t]
    \centering
    \small
    \begin{tabular}{c|c|cc|cc}
        \toprule
         & \multirow{2}{*}{\shortstack{Conv.\\history}}
        &\multicolumn{2}{c|}{DVC}&\multicolumn{2}{c}{TVG}\\
        Paths & & $\text{SODA}_c$ & METEOR & R@0.3 & mIoU \\

        \midrule
        \midrule
        \multirow{2}{*}{$\mathcal{P}_{t \rightarrow c}$} & 
        $\hat{h}_{< k}$ &
         7.8 & \textbf{8.8} & 61.3 & 42.8 \\

       & $h_{< k}$ & \textbf{8.2} & 8.7 & \textbf{63.3} & \textbf{44.1} \\

        \midrule
        \multirow{2}{*}{$\mathcal{P}_{c \rightarrow t}$} & 
        $\hat{h}_{< k}$ &
         8.3 & \textbf{8.9} & 61.3 & 42.8 \\

       & $h_{< k}$ & \textbf{8.8} & 8.8 & \textbf{63.3} & \textbf{44.1} \\
       
        \bottomrule
    \end{tabular}

    \caption{
    \textbf{Analysis on M-DPO conversation history type.}
    Note we fine-tune with VideoLLaMA2+CoTasks.
    }
    \label{supp-tab:DPO-generate}
\end{table}
\section{M-DPO data generated with predictions}
\label{sup:dpo_generate}
We here compare our conversation history type for constructing $\mathcal{D}_{\text{M-DPO}}$.
Initially, in our proposed M-DPO, we sample responses conditioned on the \textit{ground-truth} conversation history $h_{<k}$. 
Yet, we conduct experiments to compare with sample responses conditioned on the \textit{generated} conversation history, which we denote as $\hat{h}_{<k}$.
For clarification, $h_{<k}$ and $\hat{h}_{<k}$ adopt the same instruction prompt $x_{<k} $ given the sub-tasks in CoTasks for the conversation history, but $\hat{h}_{<k}$ bases on the responses $\hat{y}_{<k}$ generated from the model whereas $h_{<k}$ bases on the ground-truth responses $y_{<k}$.
Hence, we finetune VideoLLaMA2+CoTasks model with our M-DPO, but adopt modified $\mathcal{D}_{\text{M-DPO}}$.
In Tab~\ref{supp-tab:DPO-generate}, we show that constructing the training dataset with ground-truth conversation history, $h_{<k}$, overall yields reasonably higher performance than the generated ones for both reasoning path prompts $\mathcal{P}_{t \rightarrow c}$ and $\mathcal{P}_{c \rightarrow t}$.
Specifically, $\text{SODA}_c$ improves 0.4, and 0.5 for $\mathcal{P}_{t \rightarrow c}$ and $\mathcal{P}_{c \rightarrow t}$, respectively, and also in TVG, the performance gain is 2.0 and 1.3 for R@0.3 and mIoU respectively.
In terms of METEOR, $h_{<k}$ results in slightly lower performance with a decrease of 0.1, yet considering other metric evaluations, we find $h_{<k}$ to be a reasonable choice. 

\section{Inference Cost}
\label{sup:inference-cost}
\begin{table}[h!]
\centering
\begin{adjustbox}{width=0.99\columnwidth}
\begin{tabular}{@{}l|ll|l@{}}
\toprule
{Method} & {SODA\textsubscript{c}} & {CIDEr} & {Time (s)} \\ 
\toprule
VTimeLLM & 5.8 & 27.9 & 2.7 \\ 
+ VidChain (ours) & \textbf{7.2} \small{(+24\%)} & \textbf{34.7} \small{(+24\%)} & 2.8 \small{(+4\%)} \\ 
\midrule
VideoLLaMA2 & 7.1 & 32.9 & 10.7 \\ 
+ VidChain (ours) & \textbf{8.8} \small{(+24\%)} & \textbf{43.9} \small{(+33\%)} & 13.1 \small{(+22\%)} \\ 
\bottomrule

\end{tabular}
\end{adjustbox}
\caption{\textbf{Inference Time of VidChain}. Inference cost compared to the baseline model with our VidChain.}
\label{supp:inference-cost-tab}
\end{table}
In Tab~\ref{supp:inference-cost-tab}, we compare the inference cost to generate the same number of tokens with the baseline (VideoLLaMA), and after adopting our VidChain (VideoLLaMA + VidChain).
As shown, VidChain boosts performance by up to 24\% in terms of $\text{SODA}_c$, and up to 33\% in terms of CIDEr with moderate additional inference costs.

\section{Details on $\mathcal{D}_{\text{CT}}$}
\label{sup:CoTasks_dataset}
Following the best of the training protocols in VTimeLLM~\cite{huang2024vtimellm}, we construct our CoTasks training dataset $\mathcal{D}_{\text{CT}}$ with single-turn DVC samples, standard temporal video grounding (TVG) samples, standard clip-captioning samples, and our specifically designed multi-turn samples, $\mathcal{D}_{t \rightarrow c}$ and $\mathcal{D}_{c \rightarrow t}$, for DVC.
The standard temporal video grounding task aims to determine the time boundaries, \textit{i.e.}, start and end time boundaries, for the given video and event caption.
The standard clip-captioning aims to generate the event caption given the video and the event time boundaries.
The single-turn prompt is illustrated in Fig.~\ref{supp-fig:prompt}.
The TVG adopts the prompt \textit{`During which frames can we see $\langle$caption$\rangle$ in the video?'}, which is also the prompt used to evaluate the VideoLLM on the task of TVG. 
The clip-captioning adopts the prompt \textit{`Can you describe what occurred from $\langle$start$\rangle$ to $\langle$end$\rangle$ in the video?'}.
Note that each standard temporal video grounding and clip-captioning is a multi-turn QA conversation with a single objective, and we use the whole benchmark dataset to build the samples for these given tasks.
Therefore, the total size of $\mathcal{D}_{\text{CT}}$ is 50K for ActivityNet~\cite{krishna2017dense} and 6K for YouCook2~\cite{zhou2018towards}.
Also, note that we use a single instruction for each task for simplicity; however, employing a variety of instructions for each task could enhance generalizability.

\section{Benchmark Details}
\label{supp:benchmark-detail}
We evaluate our model on two DVC benchmarks, ActivityNet~\cite{krishna2017dense}, and YouCook2~\cite{zhou2018towards}.
ActivityNet Captions dataset consists of 20k videos annotated with temporally localized descriptions carefully by human annotators.
Typically the video lasts about 2 minutes, and the average number of segments for the dataset is 3.7.
For the evaluation of DVC and TVG, we adopted the val2 validation set. 
YouCook2 dataset is composed of 2,000 videos from 89 recipes.
Specifically, it focuses on cooking scenes, where one usually cooks or is a video of someone describing the recipe.
Typically the video lasts an average of 5.3 minutes, and the average number of events in the video is about 7.8 events.
We conduct experiments on the validation set.

\begin{table*}[t!]
    \centering
    \begin{adjustbox}{width=0.6\linewidth}
    \small
    \begin{tabular}{c|cc|cc}
        \toprule
         &\multicolumn{2}{c|}{VTimeLLM}&\multicolumn{2}{c}{VideoLLaMA}\\
         & {CoTasks} & M-DPO & CoTasks & M-DPO \\

        \midrule
        \midrule
        Learning rate & 2e-5 & 1e-6 & 2e-5 & 5e-7 \\
        Warmup Ratio & 0.03 & 0.1 & 0.03 & 0.1 \\
        Epoch & 1 & 1 & 1 & 1 \\
        Batch Size & 32 & 32 & 16 & 16 \\
        Accumulation Steps & 4 & 4 & 8 & 4 \\
        LoRA $r$ & 64 & 64 & 64 & 64 \\
        LoRA $\alpha$ & 128 & 128 & 128 & 128 \\
        $\beta$ & - & 0.5 & - & 0.5 \\
        $\gamma$ & - & 5.0 & - & 10.0 \\
       
        \bottomrule
    \end{tabular}
    \end{adjustbox}
    \caption{Training hyperparameters for \textbf{ActivityNet}.
    }
    \label{supp:hparams_tab}
\end{table*}

\begin{table*}[t!]
    \centering
    \begin{adjustbox}{width=0.6\linewidth}
    \small
    \begin{tabular}{c|cc|cc}
        \toprule
         &\multicolumn{2}{c|}{VTimeLLM}&\multicolumn{2}{c}{VideoLLaMA}\\
         & {CoTasks} & M-DPO & {CoTasks} & {M-DPO} \\

        \midrule
        \midrule
        Learning rate & 1e-4 & 2.5e-6 & 2e-5 & 5e-7 \\
        Warmup Ratio & 0.03 & 0.1 & 0.03 & 0.1 \\
        Epoch & 5 & 1 & 5 & 1 \\
        Batch Size & 32 & 32 & 16 & 16 \\
        Accumulation Steps & 4 & 4 & 8 & 4 \\
        LoRA $r$ & 64 & 64 & 64 & 64 \\
        LoRA $\alpha$ & 128 & 128 & 128 & 128 \\
        $\beta$ & - & 0.5 & - & 0.5 \\
        $\gamma$ & - & 10.0 & - & 10.0 \\
       
        \bottomrule
    \end{tabular}
    \end{adjustbox}
    \caption{Training hyperparameters for \textbf{YouCook2}.
    }
    \label{supp:hparams_tab_youcook}
\end{table*}

\section{Metric Details}
\label{sup:metric_detail}
\paragraph{METEOR and CIDEr for DVC.}
\label{sup:meteor_cider}
Typically METEOR and CIDEr in DVC measure the average precision with the caption evaluation metrics METEOR and CIDER across tIoU, \textit{i.e.}, temporal overlaps between the predicted and ground truth time boundaries, thresholds of 0.3, 0.5, and 0.7, suggested by \citet{krishna2017dense}.
Note that in the field of DVC, METEOR and CIDER refer to the evaluation metric that measures the ability of localization and captioning, instead of the traditional captioning evaluation metrics.
Hence, for less confusion, in this section, we denote the traditional captioning evaluation metrics METEOR~\cite{banerjee2005meteor} and CIDEr~\cite{vedantam2015cider} as $f_{\text{METEOR}}$ and $f_{\text{CIDEr}}$.
Hence, given a set of ground-truth captions for a video $C^*$, a set of predicted captions $C$, $t_c$ for the time boundaries of a given caption $c$, then $\text{METEOR} = \langle E_\text{METEOR}\left(C^*, C, \tau \right) \rangle_{\{0.3, 0.5, 0.7\}},$ where $E_\text{METEOR}$ is defined as follows: 
\begin{equation}
   E_\text{METEOR}\left(C^*, C, \tau \right) = \frac{\sum_{c^* \in C^*}\sum_{c \in \hat{C}_{c^*, \tau}}  f_{\text{METEOR}}\left(c^*, c \right)}{\sum_{c^* \in C^*} \left\lvert \hat{C}_{c^*, \tau} \right\rvert},
\end{equation}
where $\hat{C}_{c^*, \tau} = \left\{c \in C | \text{IoU}\left(t_c, t_{c^*} \right) \geq \tau \right\}$,
$\text{IoU}$ is the Intersection over Union, and $\tau$ as the thresholds for grounding.
CIDEr is also similarly defined.

\paragraph{$\text{SODA}_c$ for DVC.}
\label{sup:soda}
$\text{SODA}_c$~\cite{fujita2020soda} is a more standard evaluation metric for DVC, which evaluates how well the story of the video is described.
Hence, it evaluates under various criteria, \textit{e.g.}, the order of the generated captions, the quality of the captions, the prediction of time boundaries, and etc.
Particularly, it finds the optimal match between the generated captions $C$ and reference captions $C^*$ and seeks to maximize the total IoU while taking into account the temporal sequence of the captions.

Specifically, $\text{SODA}_c$ adopts a cost function to find the optimal matches and applies dynamic programming based on this cost function, which is defined as follows:
\begin{equation}
\Phi_{i, j} = \text{IoU}\left(c^*_i, c_j  \right)f_{\text{METEOR}}\left(c^*_i, c_j \right).
\end{equation}
Then given the optimal path, the evaluation system ensures not too many captions are generated for prediction, by measuring the harmonic mean of $\text{Precision}(C^*, C)$, and $\text{METEOR}(C^*, C)$, that is defined as follows:
\begin{equation}
    \text{Precision}\left(C^*, C \right) = \frac{\sum_{c^* \in C^*} f_{\text{METEOR}}\left(c^*, c_{\pi(c^*)}  \right)}{\left\vert C \right\rvert},
\end{equation}
\begin{equation}
    \text{Recall}\left(C^*, C \right) = \frac{\sum_{c^* \in C^*} f_{\text{METEOR}}\left(c^*, c_{\pi(c^*)}  \right)}{\left\vert C^* \right\rvert},
\end{equation}
where $\pi(\cdot)$ is the optimal assignment.
Hence, the DVC evaluation metric $\text{SODA}_c$ is formulated as:
\begin{equation}
    \text{SODA}_c\left(C^*, C \right) = \frac{2 \times \text{Precision}\left(C^*, C \right) \times \text{Recall}\left(C^*, C \right)}{\text{Precision}\left(C^*, C \right) + \text{Recall}\left(C^*, C \right)}.
\end{equation}

\paragraph{R@$k$ for TVG.}
\label{sup:recall}
R@$k$ is a performance metric used to measure how accurately the video segments are localized.
Specifically, it is defined as follows:
\begin{equation}
    R@k\left(T^*, T \right) = \frac{1}{\left\lvert T^* \right\rvert}\sum_{t^* \in T^*}{\mathbf{1}\left(\text{IoU}\left(t^*,t \right)\ge k \right)},
\end{equation}
where $T$ is the set of predicted time boundaries, $\mathbf{1}\left(\cdot\right)$ is an indicator function, and $T^*$ is the set of ground-truth time boundaries $t^*$. 

\paragraph{mIoU for TVG.}
\label{sup:mIoU}
mIoU averages the IoU scores across the predicted time intervals $T$ and the ground truth time intervals $T^*$ that is defined as follows:
\begin{equation}
    \text{mIoU}(T^*, T) = \frac{1}{\left\lvert T^* \right\rvert}\sum_{t^* \in T^*} \text{IoU}\left(t^*, t \right).
\end{equation}


\section{Training Details}
\label{sup:training_details}
Training hyperparameters for VTimeLLM and VideoLLaMA are reported in Tab.~\ref{supp:hparams_tab} and Tab.~\ref{supp:hparams_tab_youcook} for ActivityNet and YouCook2 respectively.
We used 100 frames for VTimeLLM and 32 frames for VideoLLaMA2.
Also for both training of CoTasks and M-DPO, we adopt LoRA for parameter-efficient fine-tuning, where the most time-consuming experiment for training with CoTasks was done in just 6 hours using 8 RTX A6000 GPUs, and the most time-consuming experiment for training with M-DPO was efficiently done in 9 hours.
Following VTimeLLM, after each stage of training, we merge the LORA module trained and introduce a new LORA module with the only trainable parameters.
For $\mathcal{D}_{\text{M-DPO}}$ construction, we adopt $n_s$ as 3, where we generate 3 samples for the given training set sample.
In practice, for evaluating each generated sample, we simplify the process by using the SODA metric, as SODA encompasses all evaluation metrics for sub-tasks of CoTasks.
For instance, when comparing $p(c_1|v,n)$ and $p(c_2|v,n)$, where $c_1$ and $c_2$ are the generated output captions, and $n$ represent the ground truth number of segments, the condition holds such that using SODA is equivalent to evaluating with the captioning metric METEOR.
Overall, after filtering with the threshold for the gap of the evaluation metric, $\gamma$, the pairwise samples for ActivityNet are left with 20k and 40k for VideoLLaMA and VTimeLLM respectively.
Similarly, the pairwise samples for YouCook2 are left with 5k and 1k.
Note that increasing the sampling rate can result in greater samples for each dataset, potentially enhancing the performance of the model.
Also, we mention that there are a few missing videos that are no longer available online for both datasets.

\section{Further qualitative examples}
\label{sup:qualitatives}
\begin{figure*}[t!]
    \centering
    \includegraphics[width=1\textwidth]{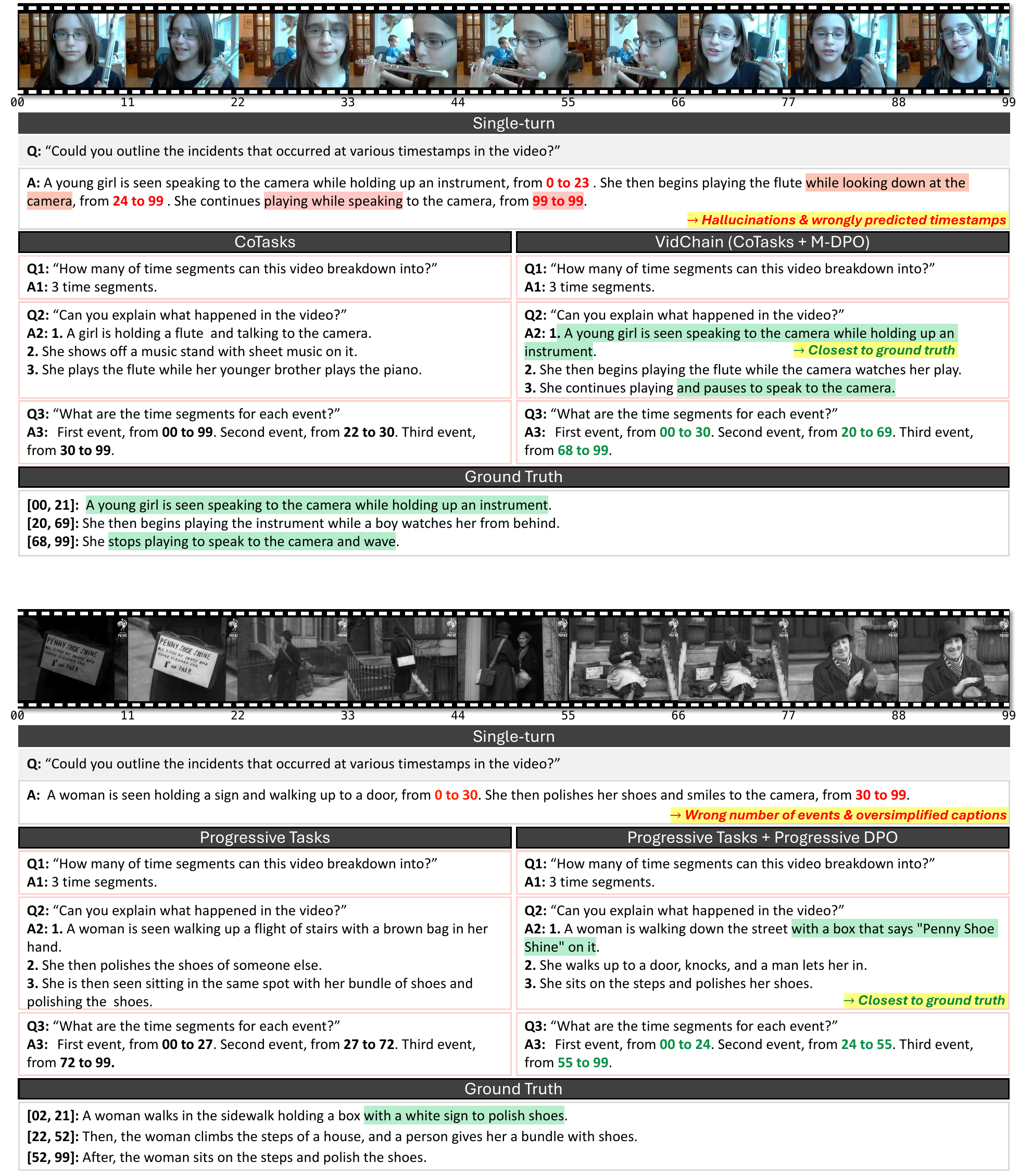}
    \caption{\textbf{Qualitative examples of DVC prediction with VideoLLaMA2 on ActivityNet.}} 
    \label{supp-fig:videollama-act1}
\end{figure*}

\begin{figure*}[t!]
    \centering
    \includegraphics[width=1\textwidth]{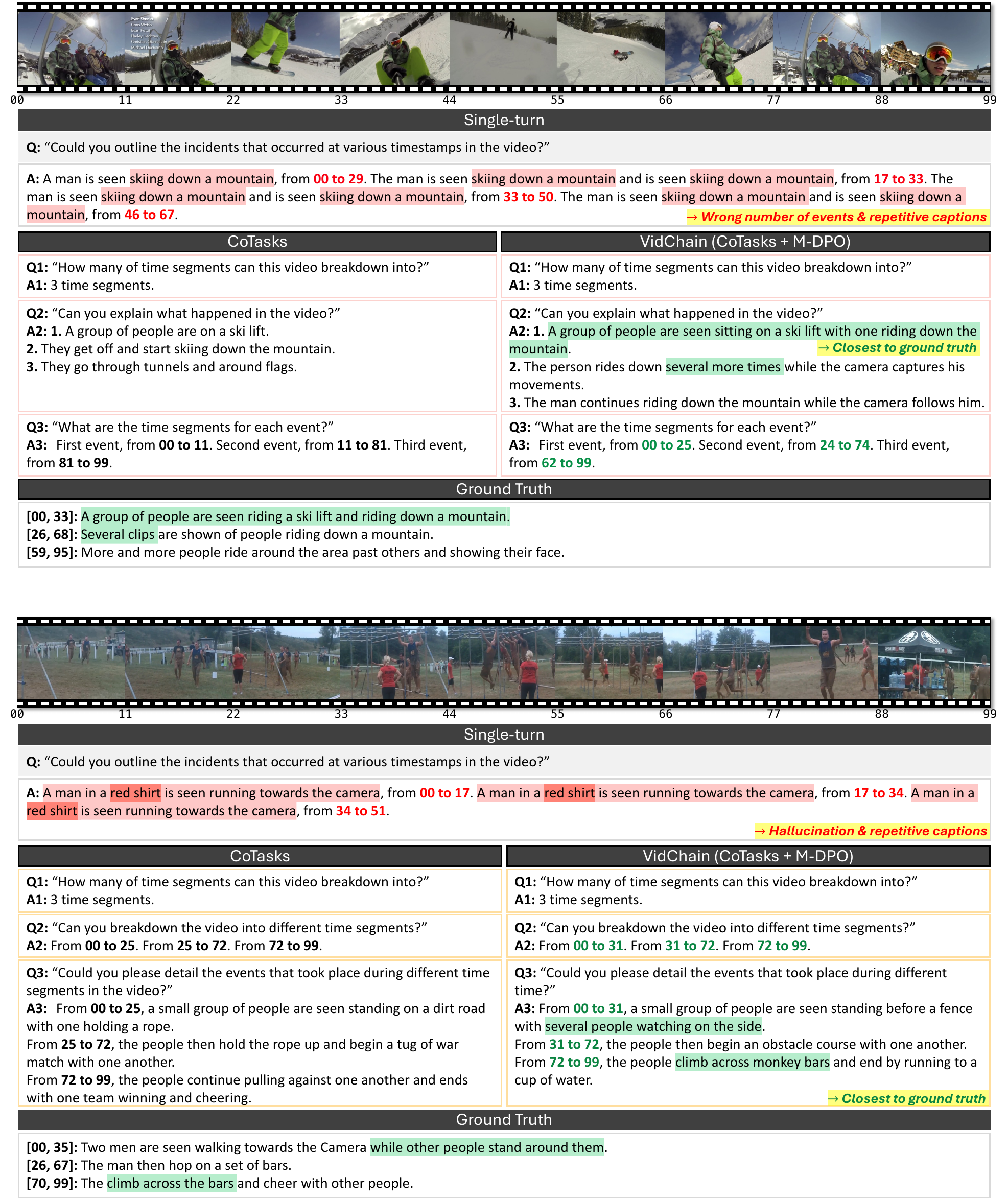}
    \caption{\textbf{Qualitative examples of DVC prediction with VTimeLLM on ActivityNet.}} 
    \label{supp-fig:vtimellm-act1}
\end{figure*}

\begin{figure*}[!ht]
    \centering
    \includegraphics[width=0.95\textwidth]{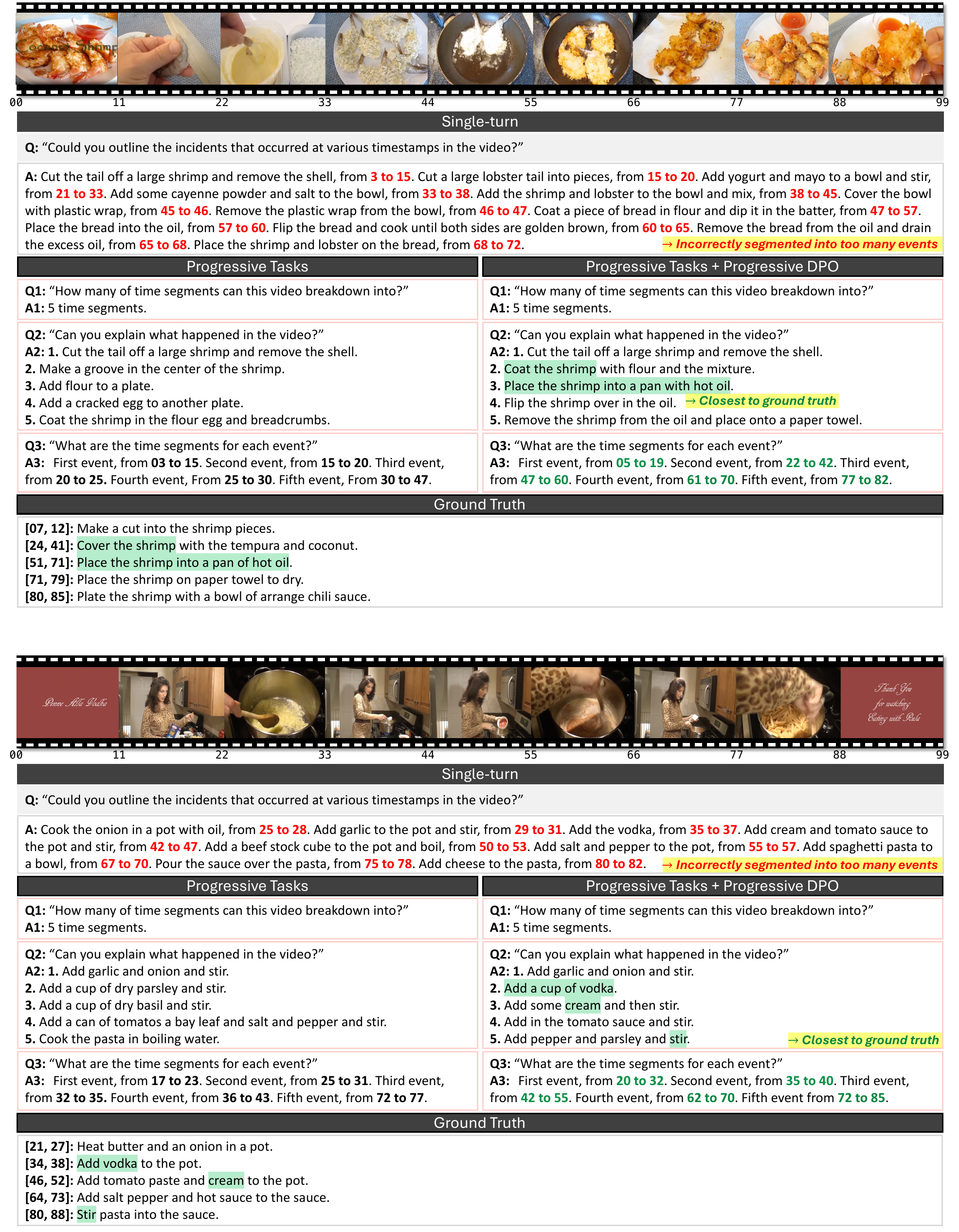}
    \caption{\textbf{Qualitative examples of DVC prediction with VTimeLLM on YouCook2 with inference prompt $\mathcal{P}_{c \rightarrow t}$.}} 
    \label{supp-fig:vtimellm-youcook1}
\end{figure*}

\begin{figure*}[t!]
    \centering
    \includegraphics[width=1\textwidth]{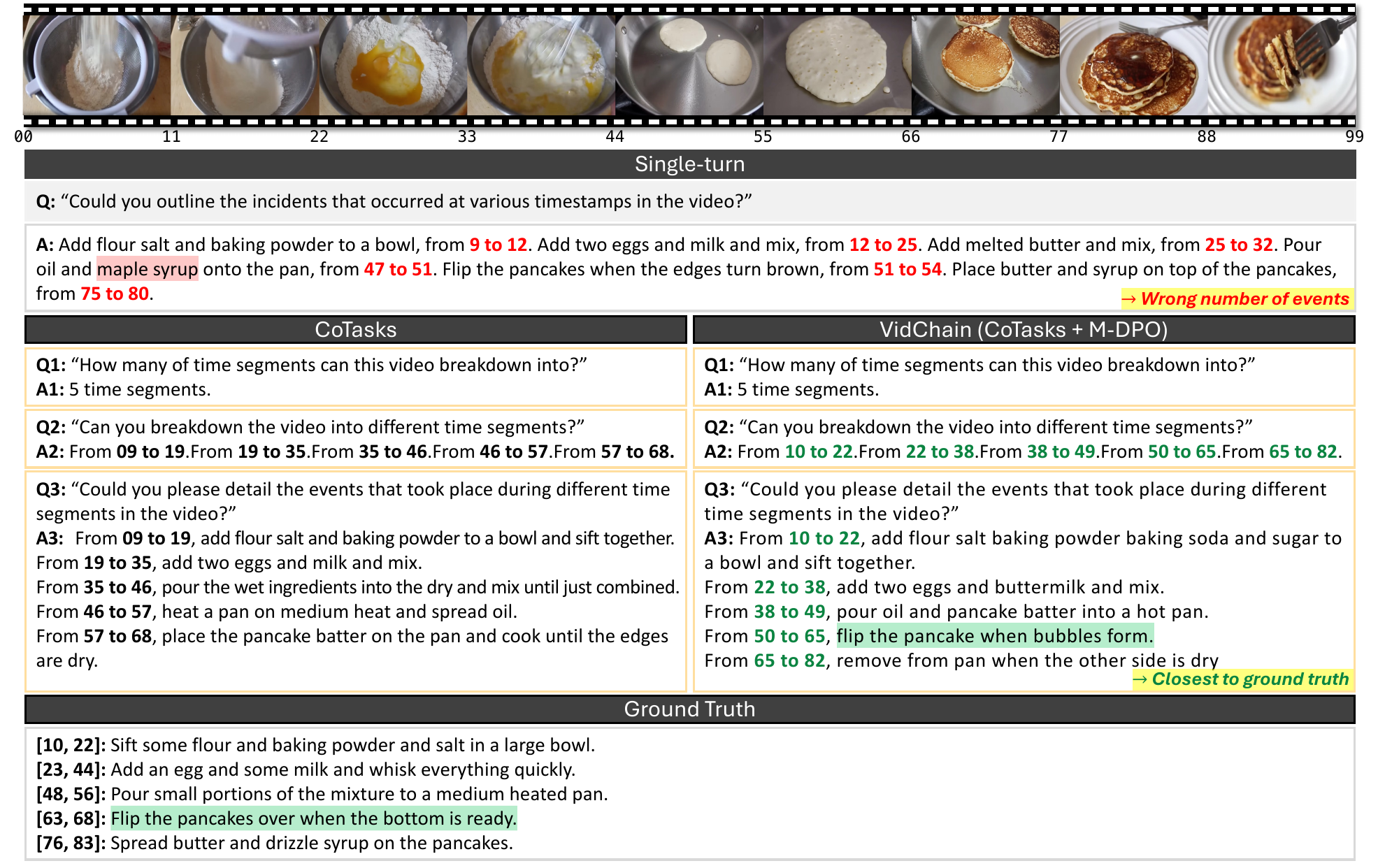}
    \caption{\textbf{Qualitative examples of DVC prediction with VTimeLLM on YouCook2 with inference prompt $\mathcal{P}_{t \rightarrow c}$.}} 
    \label{supp-fig:vtimellm-youcook2}
\end{figure*}

We provide more qualitative examples of our VidChain.
DVC prediction result with VideoLLaMA2 on AcitivityNet is illustrated in Fig.~\ref{supp-fig:videollama-act1}.
For DVC prediction with VTimeLLM, Fig.~\ref{supp-fig:vtimellm-act1} illustrates the predictions on ActivityNet, Fig.~\ref{supp-fig:vtimellm-youcook1}, Fig.~\ref{supp-fig:vtimellm-youcook2} illustrates predictions with the inference prompt $\mathcal{P}_{c \rightarrow t}$, and $\mathcal{P}_{t \rightarrow c}$ on YouCook2, respectively.


\end{document}